\RequirePackage[T1]{fontenc}
\documentclass[letterpaper,journal]{IEEEtran}
\usepackage{amsmath,amssymb,graphicx,cite,booktabs,multirow}
\usepackage{newtxtext,newtxmath}
\usepackage[caption=false,font=footnotesize]{subfig}
\usepackage{tikz,pgfplots}
\pgfplotsset{compat=1.18}
\usepackage{url}
\newcommand{\reconpanel}[3]{\begin{tikzpicture}
\node[inner sep=0,anchor=south west] (im) at (0,0) {\includegraphics[width=#2,height=#3,trim=0bp 0bp 275bp 0bp,clip]{#1}};
\begin{scope}[x={(im.south east)},y={(im.north west)}]
\fill[white] (.965,0) rectangle (1,.916);
\end{scope}\end{tikzpicture}}
\newcommand{\gatherpanel}[2]{\ifnum#2=0\includegraphics[width=.139\textwidth,height=1.17in]{#1}\else\reconpanel{#1}{.139\textwidth}{1.17in}\fi}
\begin{document}
\title{Spatial-Spectral Interactive POCS Network for Large-Gap Seismic Data Interpolation}
\author{Baisong~Jiang, Chunxia~Zhang, Anxiang Di, Xiaoli Wei, Xiangyu Zhao, Deng Xiong, Chuangji Meng
	\thanks{Corresponding author: Chunxia Zhang. E-mail: cxzhang@mail.xjtu.edu.cn.}
	\thanks{Baisong Jiang, Chunxia Zhang, Anxiang Di, Xiaoli Wei, Xiangyu Zhao are with the School of Mathematics and Statistics, Xi'an Jiaotong University, Xi'an, Shaanxi, 710049, China.}
	\thanks{Chuangji Meng is with the School of Information and Communications Engineering, Xi’an Jiaotong University, Xi’an, Shaanxi, 710049, China.}
	\thanks{Deng Xiong is with the Geophysical Technology Research and Development Center, BGP, Zhuozhou, Hebei, 072751, China}
	\thanks{The research is supported by the National Natural Science Foundation of China under grant 12371512.}
}

\markboth{IEEE Geoscience and Remote Sensing Letters}{Jiang \MakeLowercase{\textit{et al.}}: Spatial-Spectral Interactive POCS Network}
\maketitle
\begin{abstract}
Large gaps in seismic acquisition disrupt reflection events and hinder spectral recovery. Existing seismic interpolation deep unfolding networks primarily rely on localized spatial priors, limiting their ability to aggregate the global context necessary for reconstructing these extensive missing regions. We propose a spatial-spectral interactive projection-onto-convex-sets network (SSI-POCS) to tackle this severely ill-posed problem. By unfolding the classical iterative solver, our network integrates physical consistency with dual-domain deep priors via learned transforms, soft thresholding, and exact data reinsertion. Within the learned prior, a spatial Restormer branch captures local waveform structures, while a fast Fourier convolution (FFC) branch extracts global context, effectively suppressing incoherent components and enhancing spectral energy concentration in the $f-k$ domain. A dual-domain fusion module exchanges these complementary features via bidirectional channel attention and gated modulation, yielding substantial interpolation improvements. To overcome the severe loss of lateral continuity caused by large gaps, we introduce structural similarity (SSIM) supervision, complemented by focal frequency loss (FFL) and transform consistency constraints to suppress incoherent high-frequency components and regularize training. Experiments on a synthetic salt dome and a North Sea field dataset under 50\% random and consecutive trace removal demonstrate the effectiveness of SSI-POCS. It significantly enhances event continuity across large gaps and mitigates artifacts, achieving a 3.285~dB SNR improvement over the best competing method while reducing parameters by 75.66\%.
\end{abstract}

\section{Introduction}
\IEEEPARstart{A}{cquiring} complete seismic data is essential for subsurface imaging and reservoir characterization. Large contiguous gaps commonly arise from terrain obstacles, receiver failures, and acquisition restrictions. Unlike random missing traces, a wide gap provides limited observations at its center, turning interpolation into long-range event extrapolation. The slopes and amplitudes of reflection events become increasingly difficult to constrain, often leading to broken events, weak-signal loss, and coherent artifacts, particularly in the $f$-$k$ domain.

Traditional interpolation methods impose wave-equation, prediction-filtering, or sparse-transform priors \cite{1987Ronen,1991spitz,2007Zwartjes,2017Zhang,2018Gong}. Although effective for moderate or dispersed missing traces, their performance depends on assumptions such as accurate physical models, local stationarity, or favorable sampling conditions. Their performance deteriorates when the gap exceeds the effective support of the prior, and Curvelet-POCS has been reported to struggle with overlarge gaps \cite{Hennenfent2008Gap,Yang2012Curvelet}.

Data-driven networks learn nonlinear event representations for consecutive missing traces \cite{Yu2022Attention,Huang2022SelfSupervised,Wei2022Hybrid,Cheng2024SIT,Tian2025FRUNet}. However, ordinary end-to-end networks do not inherently impose a hard data-fidelity constraint. Under the severe ill-posedness of a large gap, they may generate plausible but unsupported events, while standard end-to-end formulations do not explicitly guarantee exact preservation of the acquired traces.

POCS provides a structured iterative framework that separates prior-based reconstruction from explicit acquisition constraints. Each iteration estimates the missing region using a prescribed prior and then projects the estimate onto the acquired-data constraint set, enabling iterative refinement while preserving the measured traces. Deep unfolding retains this algorithmic structure while replacing handcrafted priors with learnable nonlinear transforms \cite{2023Chen,2024Wu,2026Wang}. However, existing seismic DUNs predominantly employ local or multiscale spatial priors, which may provide insufficient global context for connecting the two boundaries of a wide gap. This motivates SSI-POCS, which preserves the iterative constraint mechanism of POCS while introducing complementary local spatial and global spectral priors through explicit spatial--spectral interaction. 

The main contributions are:
1) Physics-Guided Unrolling Framework: We formulate the interpolation as a fixed-stage SSI-POCS solver. It combines a learned dual-domain prior with unrolled exact data reinsertion, strictly preserving the acquired traces while progressively refining the large missing region.
2) Spatial-Spectral Interactive Prior: We couple a spatial Restormer branch with a spectral fast Fourier convolution (FFC) branch via a gated dual-domain fusion module (DFM). This explicit interaction enhances global contextual modeling, which is beneficial for mitigating the severe ill-posedness of large gaps.
3) Multi-Dimensional Supervision: We complement amplitude fidelity with structural similarity (SSIM)~\cite{wang2004imagequa} and focal frequency loss (FFL)~\cite{2021jiang}. SSIM promotes structural consistency and lateral event continuity across large consecutive gaps, while FFL effectively suppresses incoherent high-frequency components in shallow blank zones.

Comprehensive experiments demonstrate the effectiveness of SSI-POCS under both large contiguous gaps and random undersampling, together with a favorable reconstruction-cost trade-off.
\begin{figure}[htp]
	\centering
	\hspace{-8mm}
	\includegraphics[width=1\columnwidth]{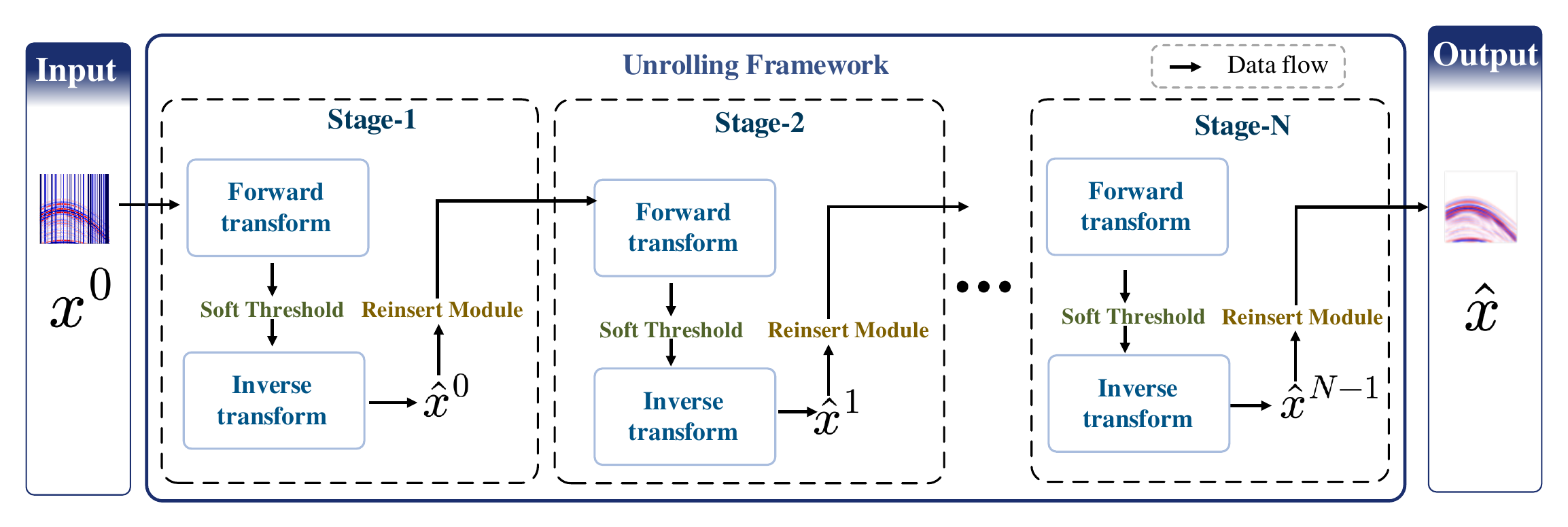}
	\caption{Overall architecture of the proposed deep unrolling framework.}
	\label{fig:workflow}
\end{figure}

\section{Methodology}\label{Methodology}
\subsection{Problem Formulation and Unfolding Framework}
Let $x\in\mathbb R^{H\times W}$ be complete seismic data, with time and trace dimensions $H$ and $W$. The available input is $y=M\odot x$, where $M$ equals one at acquired samples and zero elsewhere, and $\odot$ denotes elementwise multiplication. Interpolation seeks a complete estimate $\hat x$ that recovers event structure and spectral content inside the gaps while satisfying $M\odot\hat x=y$. A classical sparse model expresses this objective as
\begin{equation}
 \min_{\hat x}\|\Psi {\hat x}\|_1\quad\text{subject to}\quad M\odot {\hat x}=y,\label{eq:problem}
\end{equation}
where $\Psi$ is a prescribed sparsifying transform. Wide gaps weaken local constraints, motivating a learned prior that combines spatial structure with global spectral context.

As illustrated in Fig. \ref{fig:workflow}, SSI-POCS initializes with $x_0=y$ and unfolds the iterative reconstruction into $N$ stages. Within each stage, the data flow sequentially passes through a forward transform, soft thresholding, and an inverse transform, followed by a data reinsertion module that explicitly enforces measurement consistency before passing the estimate to the next step. Ultimately, the network yields the final reconstructed wavefield $\hat{x}=x^N$. Functioning as a learnable POCS-inspired solver, our framework strictly honors the hard data sampling constraints, while replacing the fixed sparse prior in Eq. \eqref{eq:problem} with powerful, stage-specific deep nonlinear transforms.

\subsection{Single-Stage Spatial--Spectral Reconstruction}

\begin{figure}[htp]
	\centering
	\hspace{-2mm}
	\includegraphics[width=1\columnwidth]{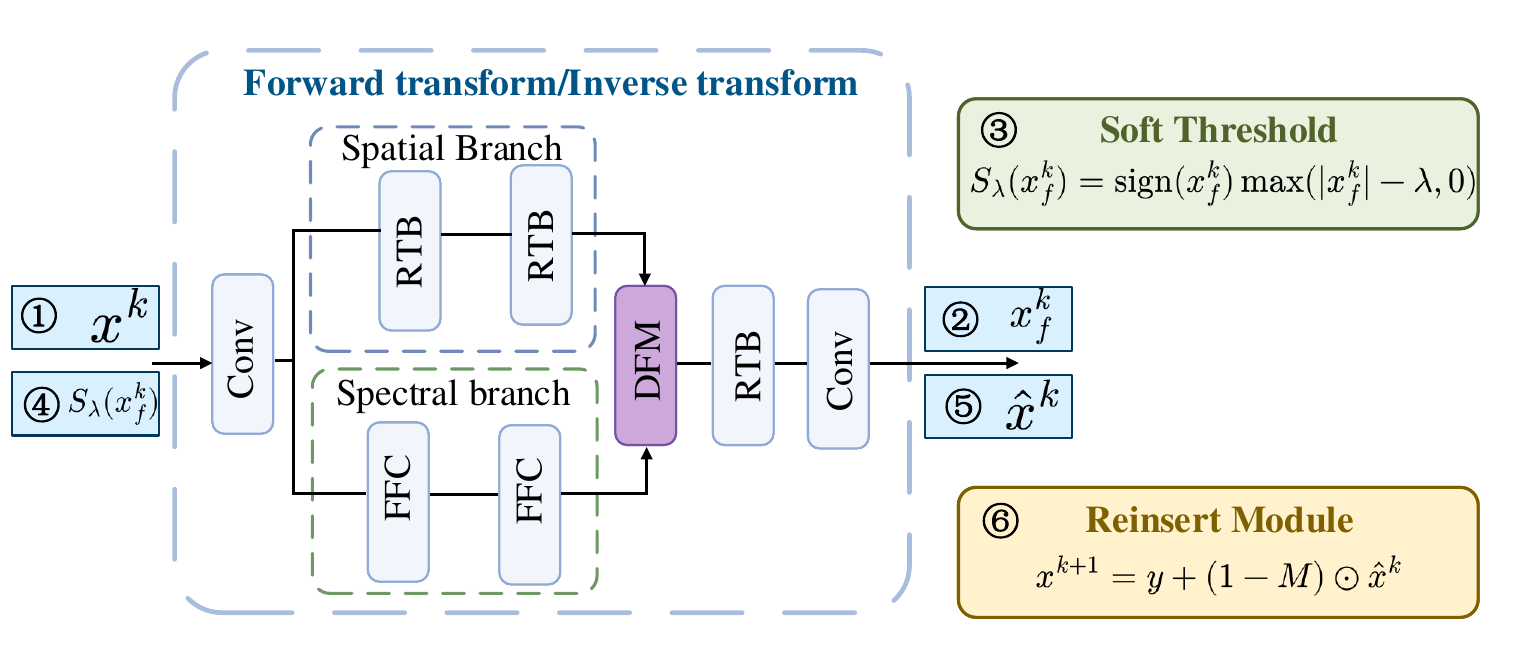}
	\caption{Detailed architecture of a single unrolled stage.}
	\label{fig:models}
\end{figure}

Fig.~\ref{fig:models} details the internal architecture of a single unrolled stage, comprising forward and inverse transforms with shared layouts but independent parameters. Following the data flow (\textcircled{1}--\textcircled{2}), the input estimate $x^k$ ($k=0,\ldots,N$) is first embedded into $C$ channels. Subsequently, a heterogeneous dual-branch prior extracts complementary features: the upper spatial branch employs Restormer blocks (RTBs)~\cite{2022Restormer} to capture local waveform structures, while the lower spectral branch utilizes FFC blocks~\cite{chi2020fast} to extract global context across missing traces. These features are dynamically integrated via the DFM to yield the dual-domain representation $x_f^k=T_k(x^k)$. 

Next (\textcircled{3}), a soft thresholding operator imposes sparsity on the learned representation:
\begin{equation}
	S_{\lambda_k}(x_f^k)=\operatorname{sign}(x_f^k)\odot\max(|x_f^k|-\lambda_k,0),\label{eq:threshold}
\end{equation}
where $\lambda_k\ge0$ is a learnable stage-specific threshold. The sparsified features $S_{\lambda_k}(x_f^k)$ are then fed into the inverse transform (\textcircled{4}--\textcircled{5}). Mirroring the forward architecture, it repeats the spatial--spectral extraction and maps the features back to the seismic-amplitude space to produce $\hat{x}^k$. To ensure stable forward propagation, a learnable convex relaxation dynamically balances $\hat{x}^k$ with the previous estimate $x^k$ via a residual connection. 

Finally (\textcircled{6}), the exact data reinsertion module strictly enforces acquisition consistency:
\begin{equation}
	x^{k+1}=y+(1-M)\odot \hat{x}^k.\label{eq:consistency}
\end{equation}

Equation~\eqref{eq:consistency} ensures strict measurement fidelity while learnable transforms replace the fixed sparse prior.
\begin{figure}[htp]
	\centering
	\hspace{0mm}
	\includegraphics[width=1\columnwidth]{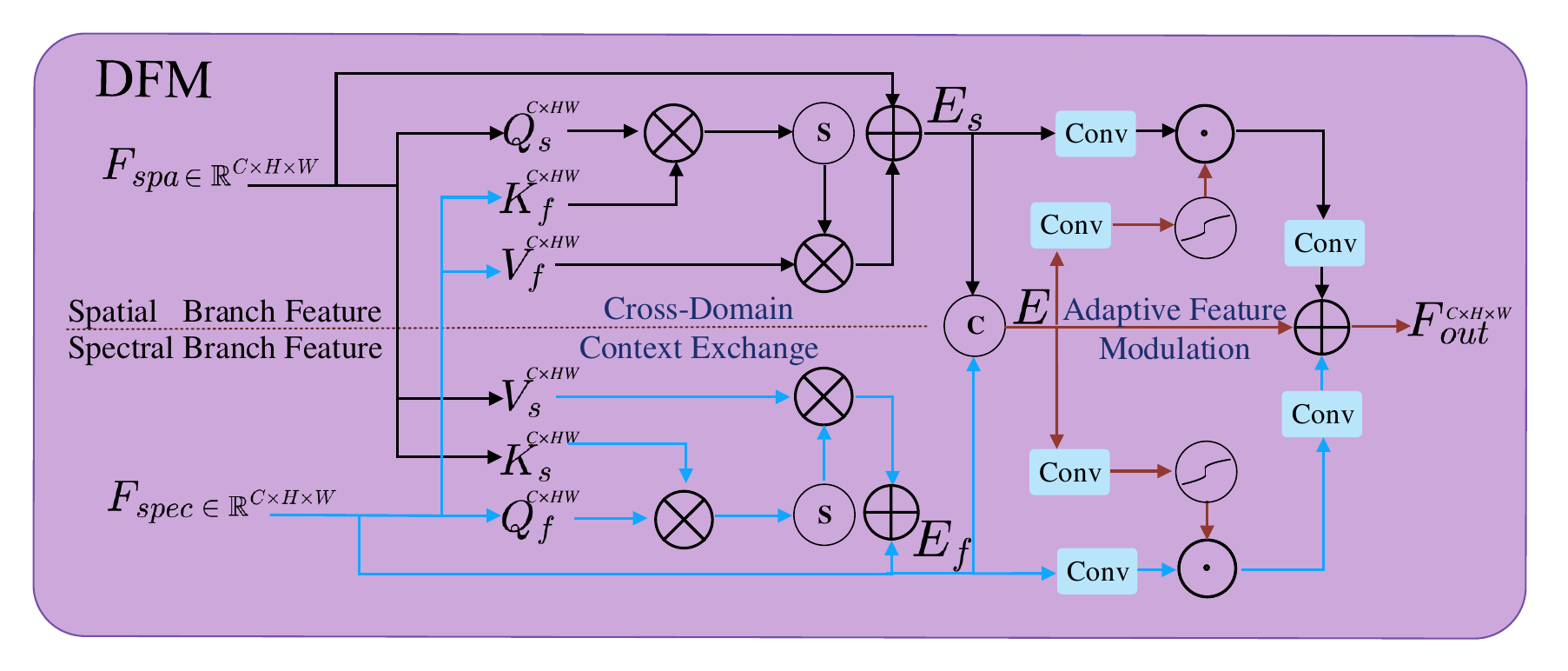}
	\caption{Detailed structure of the Dual-domain Fusion Module (DFM).}
	\label{fig:DFM}
\end{figure}

\subsection{Dual-Domain Fusion Module}
Fig.~\ref{fig:DFM} details the Dual-domain Fusion Module (DFM), which sequentially performs cross-domain context exchange and adaptive feature modulation. Given spatial ($F_{spa}$) and spectral ($F_{spec}$) features, pointwise and depthwise convolutions generate queries ($Q$), keys ($K$), and values ($V$) for each branch. To bridge the domain gap, DFM employs bidirectional cross-channel attention~\cite{2025bai}: the spatial branch utilizes $Q_s$ to aggregate spectral context from $K_f$ and $V_f$, while the spectral branch symmetrically queries $K_s$ and $V_s$. The cross-attended features are residually added to their respective inputs, yielding enhanced representations $E_s$ and $E_f$.

Subsequently, the concatenated feature $E=[E_s,E_f]$ enters the modulation stage. Independent convolutions followed by GELU activations generate dynamic gating weights. These gates selectively modulate the branch-specific features via element-wise multiplication. Finally, the modulated features are aggregated and residually added to the identity concatenation path $E$ to produce $F_{\mathrm{out}}$. This design explicitly overcomes the limitations of naive concatenation by adaptively fusing local structures with global spectral contexts.

\begin{figure*}[t]
	\centering
	\makebox[.139\textwidth]{\footnotesize Reference}\hfill\makebox[.139\textwidth]{\footnotesize Missing}\hfill\makebox[.139\textwidth]{\footnotesize Curvelet}\hfill\makebox[.139\textwidth]{\footnotesize POCSNet}\hfill\makebox[.139\textwidth]{\footnotesize SPNet}\hfill\makebox[.139\textwidth]{\footnotesize D3UNet}\hfill\makebox[.139\textwidth]{\footnotesize SSI-POCS}\par
	\subfloat[MAVO, random (shot 180)]{\label{fig:mavo_random_results}\begin{minipage}{\textwidth}\centering
			\gatherpanel{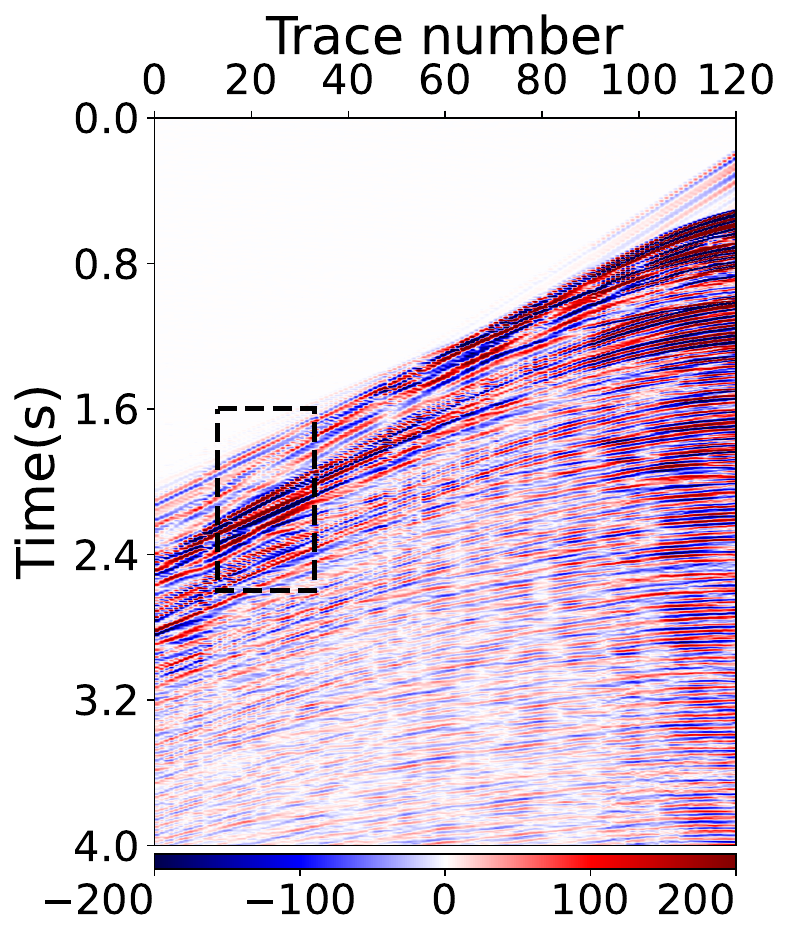}{0}\hfill
			\gatherpanel{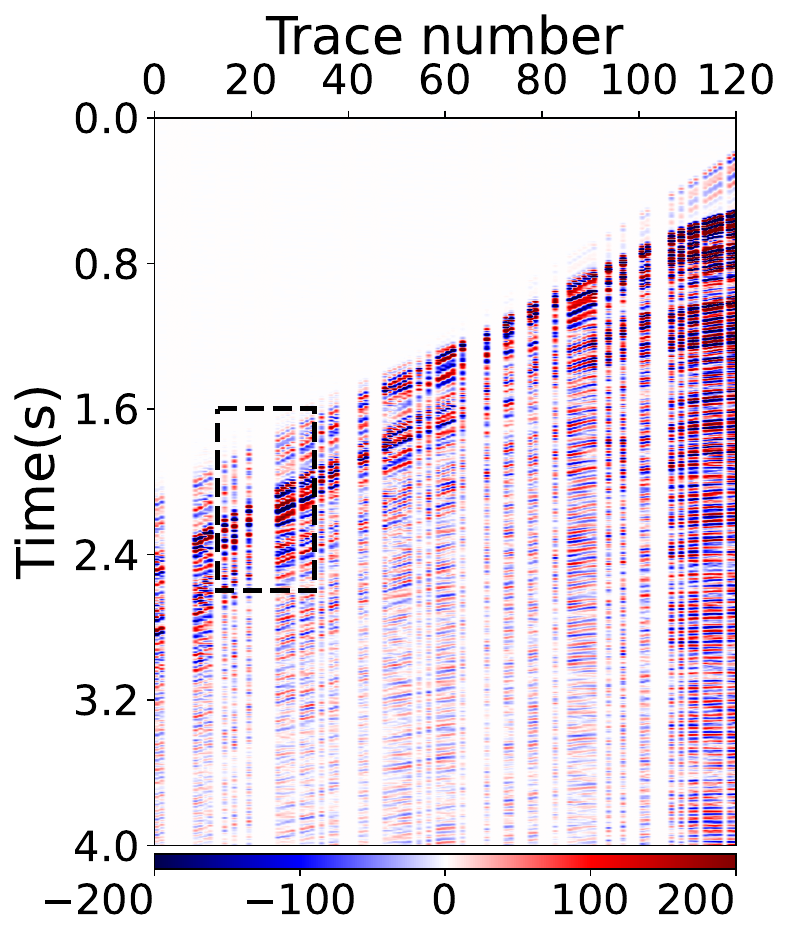}{0}\hfill
			\gatherpanel{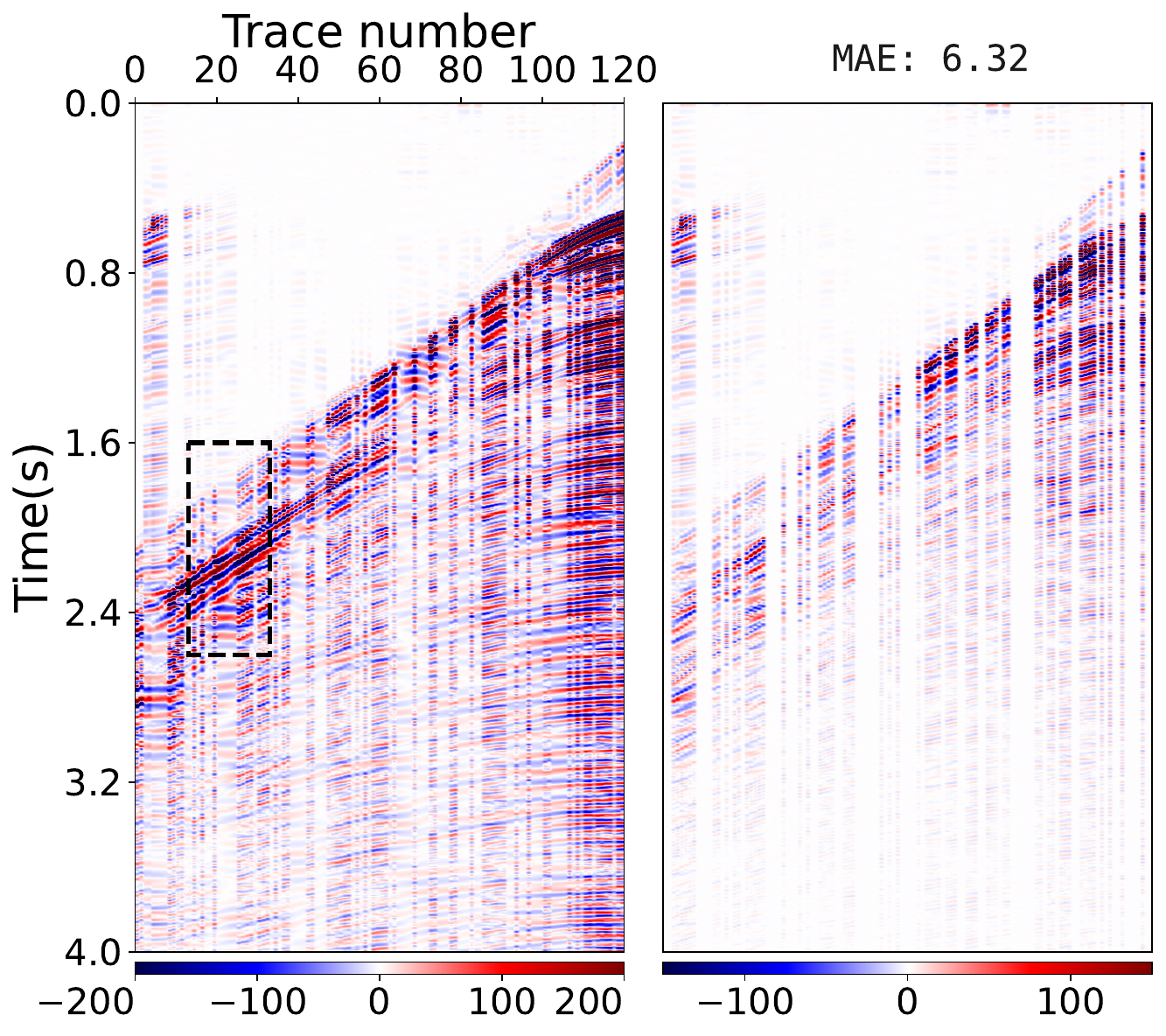}{1}\hfill
			\gatherpanel{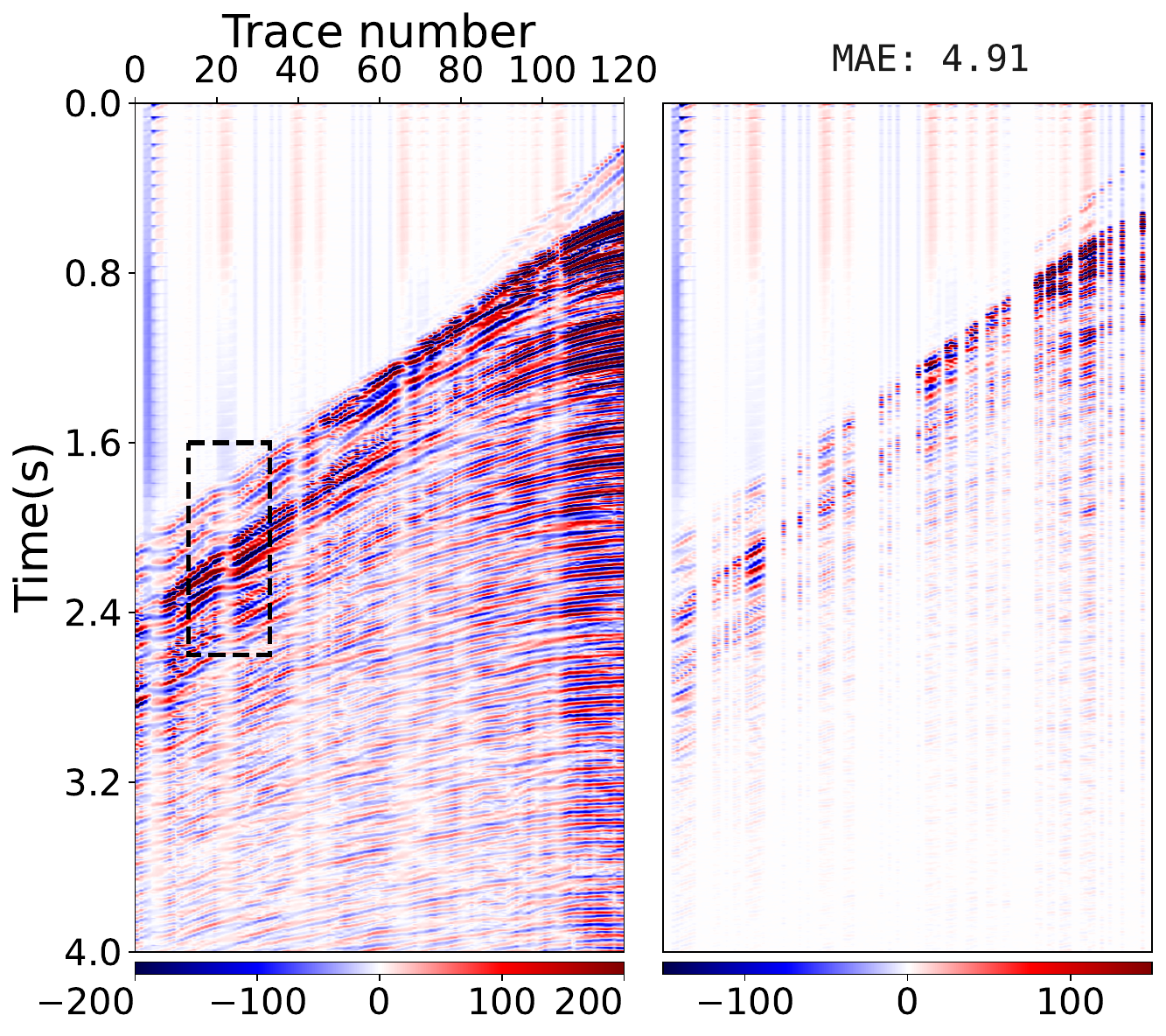}{1}\hfill
			\gatherpanel{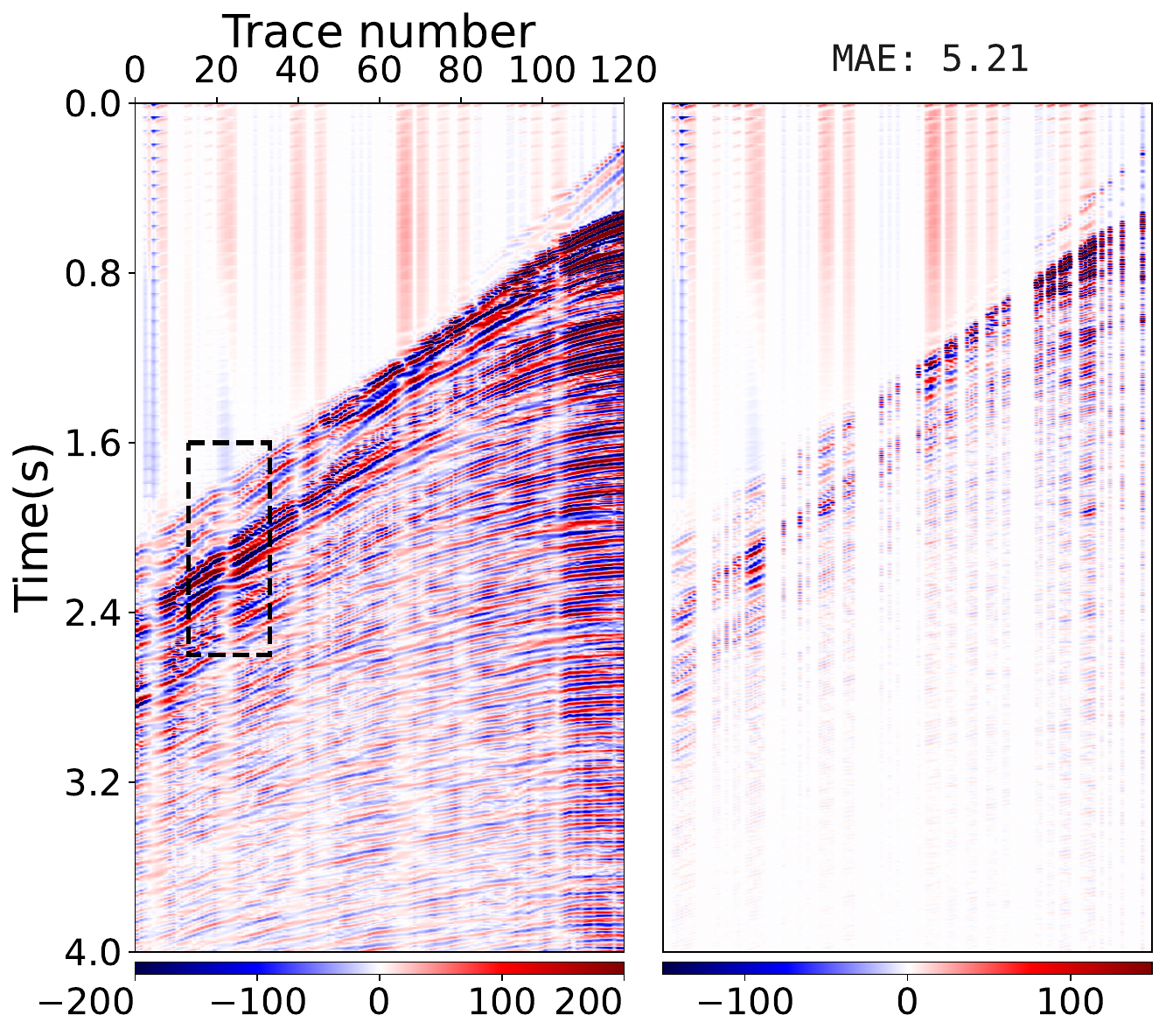}{1}\hfill
			\gatherpanel{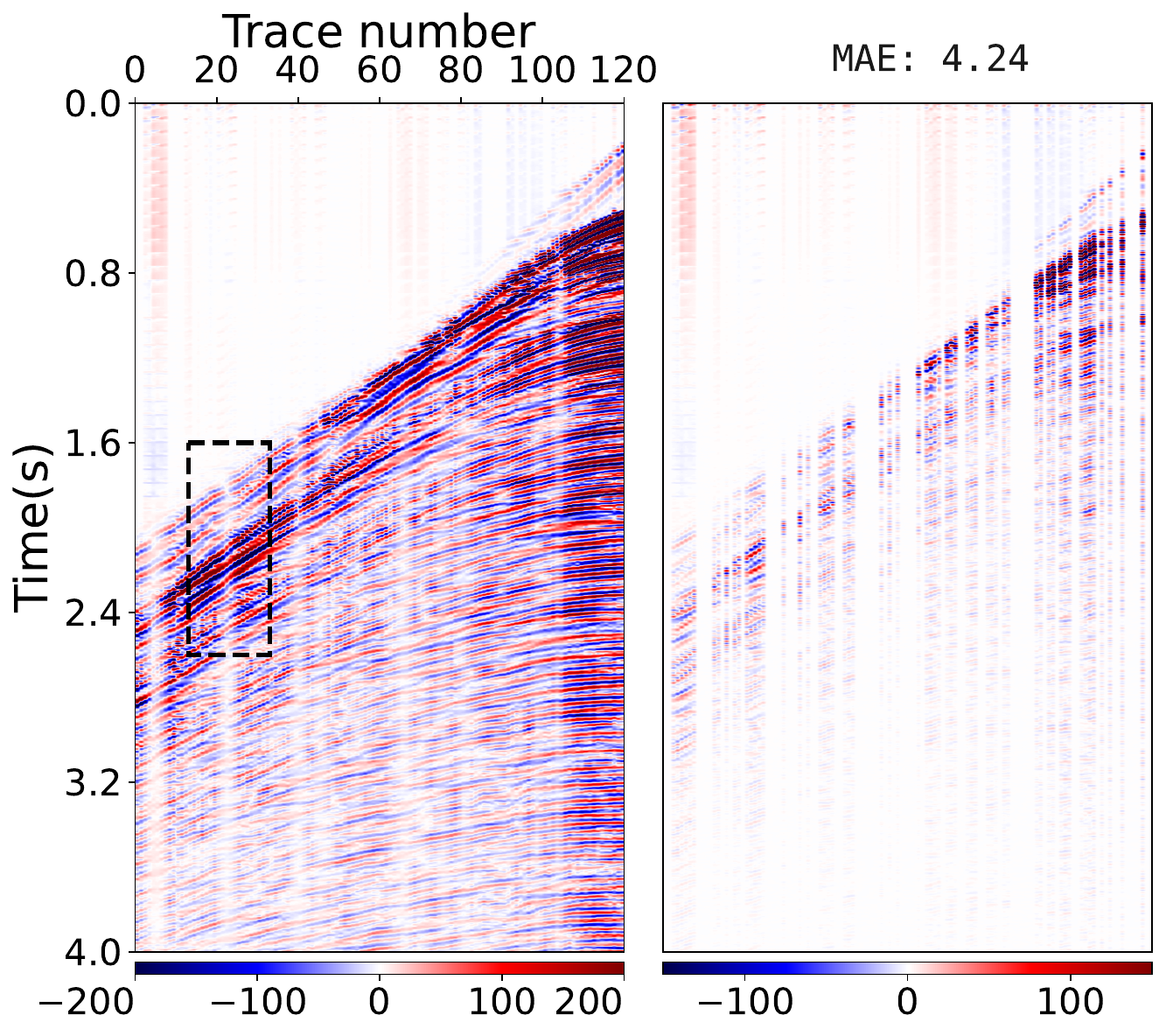}{1}\hfill
			\gatherpanel{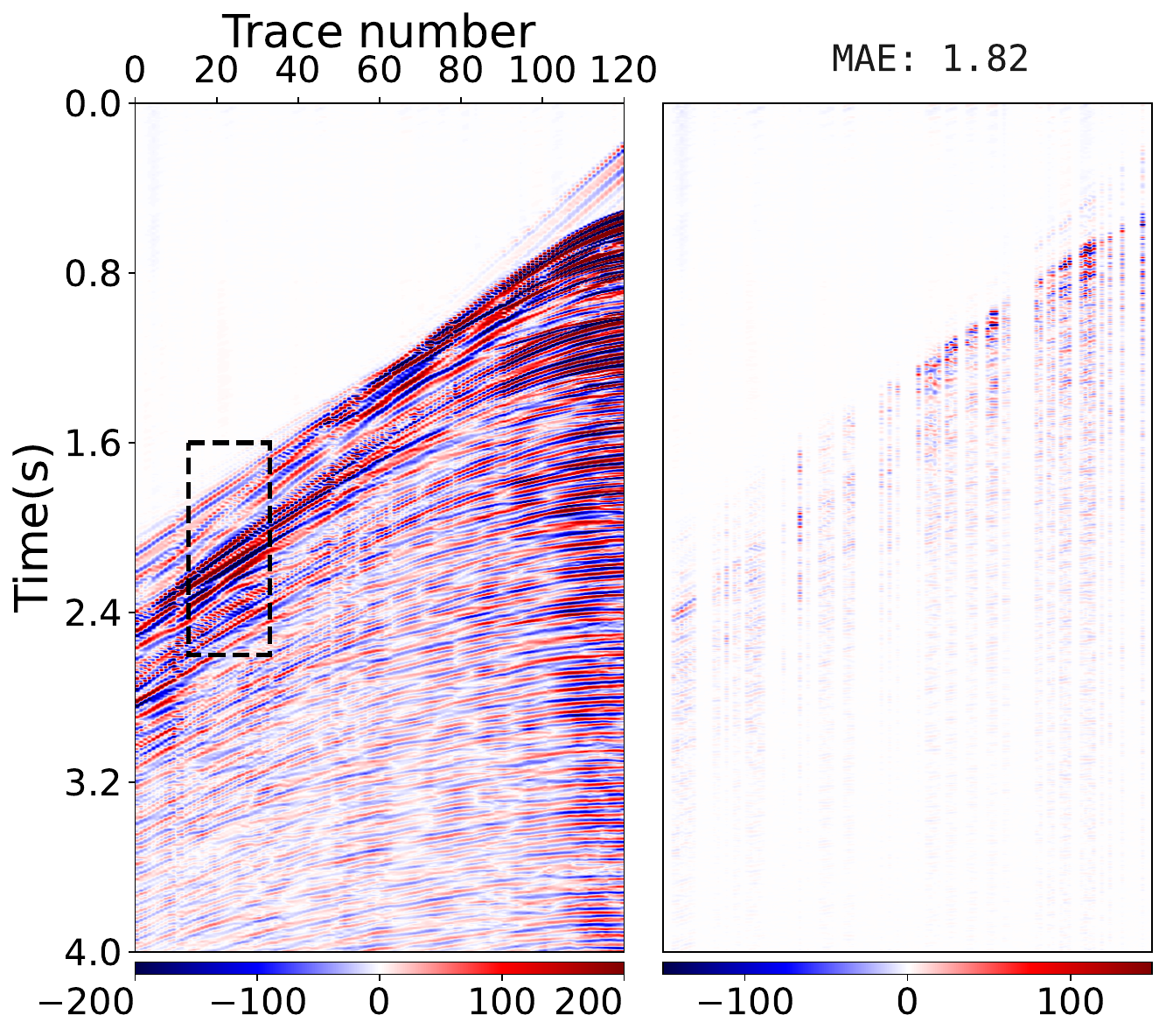}{1}
	\end{minipage}}\par
	
	\subfloat[MAVO, large gap (shot 548)]{\label{fig:mavo_continus_results}\begin{minipage}{\textwidth}\centering
			\gatherpanel{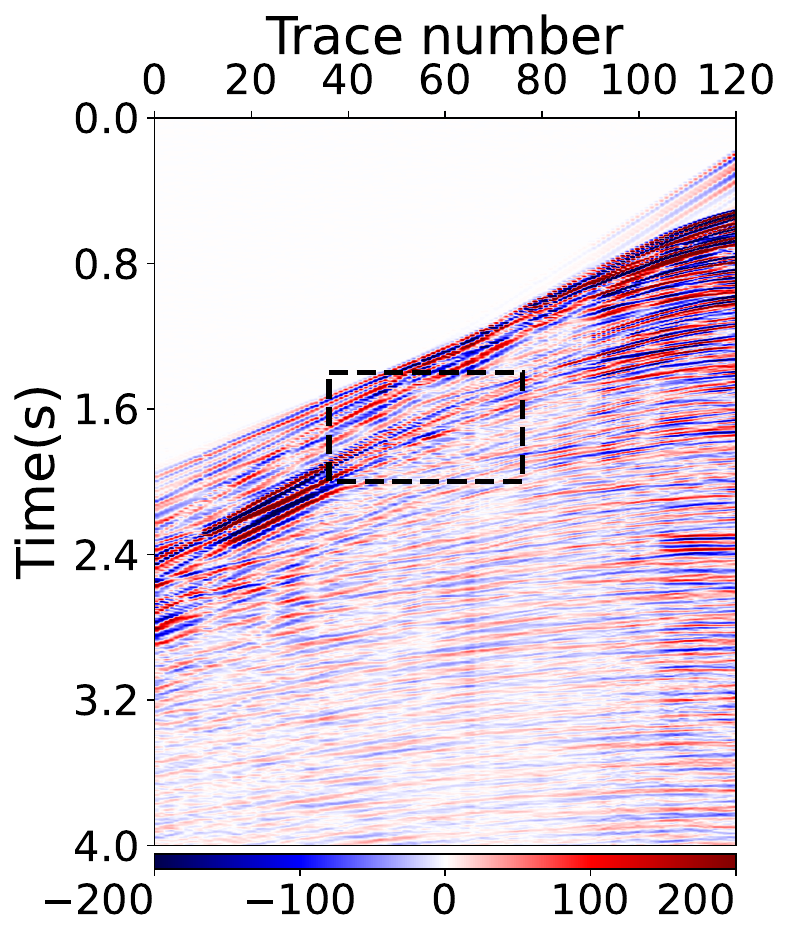}{0}\hfill
			\gatherpanel{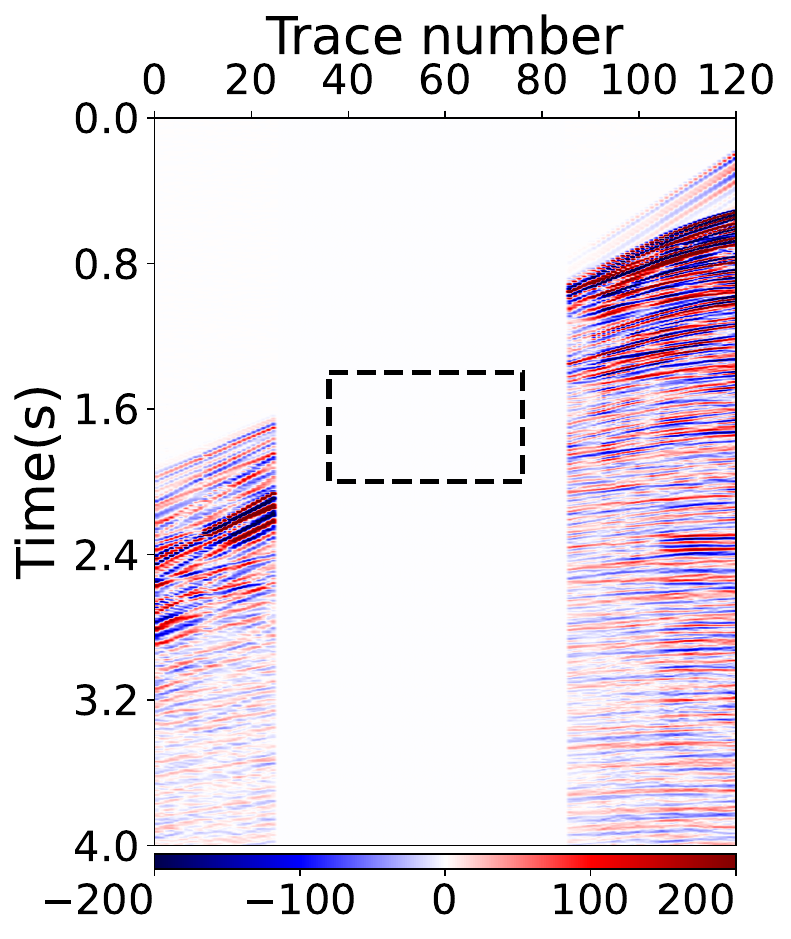}{0}\hfill
			\gatherpanel{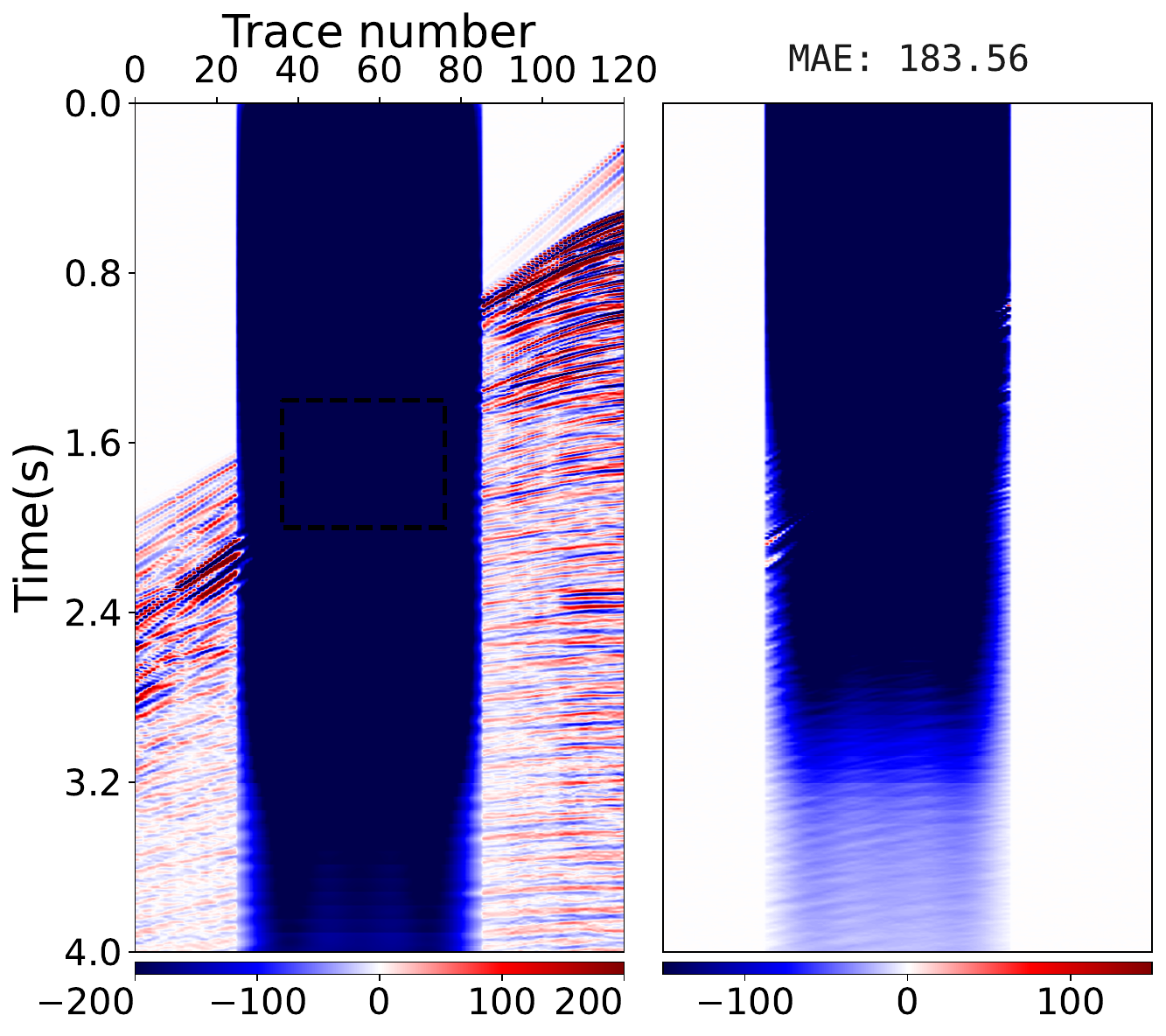}{1}\hfill
			\gatherpanel{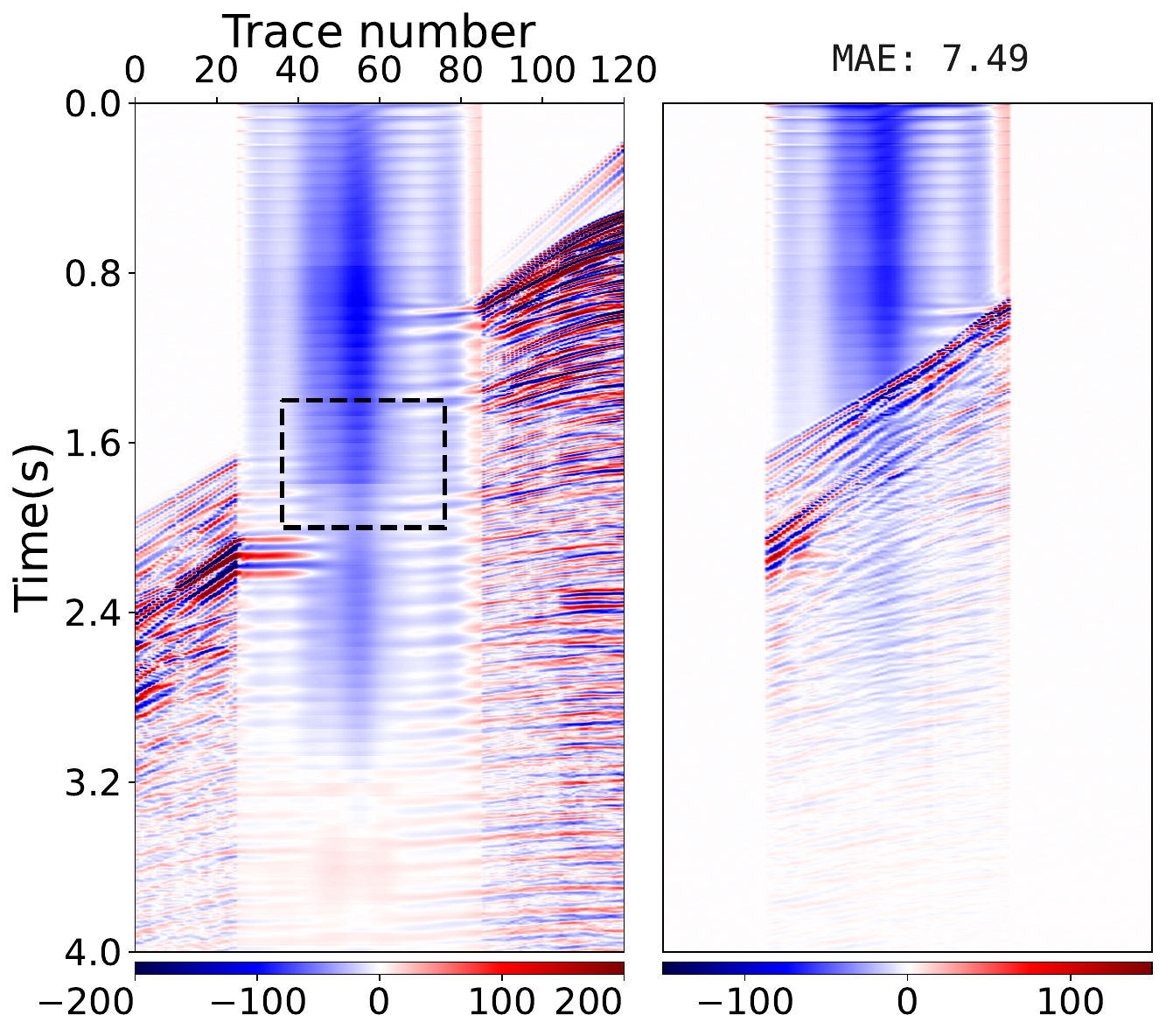}{1}\hfill
			\gatherpanel{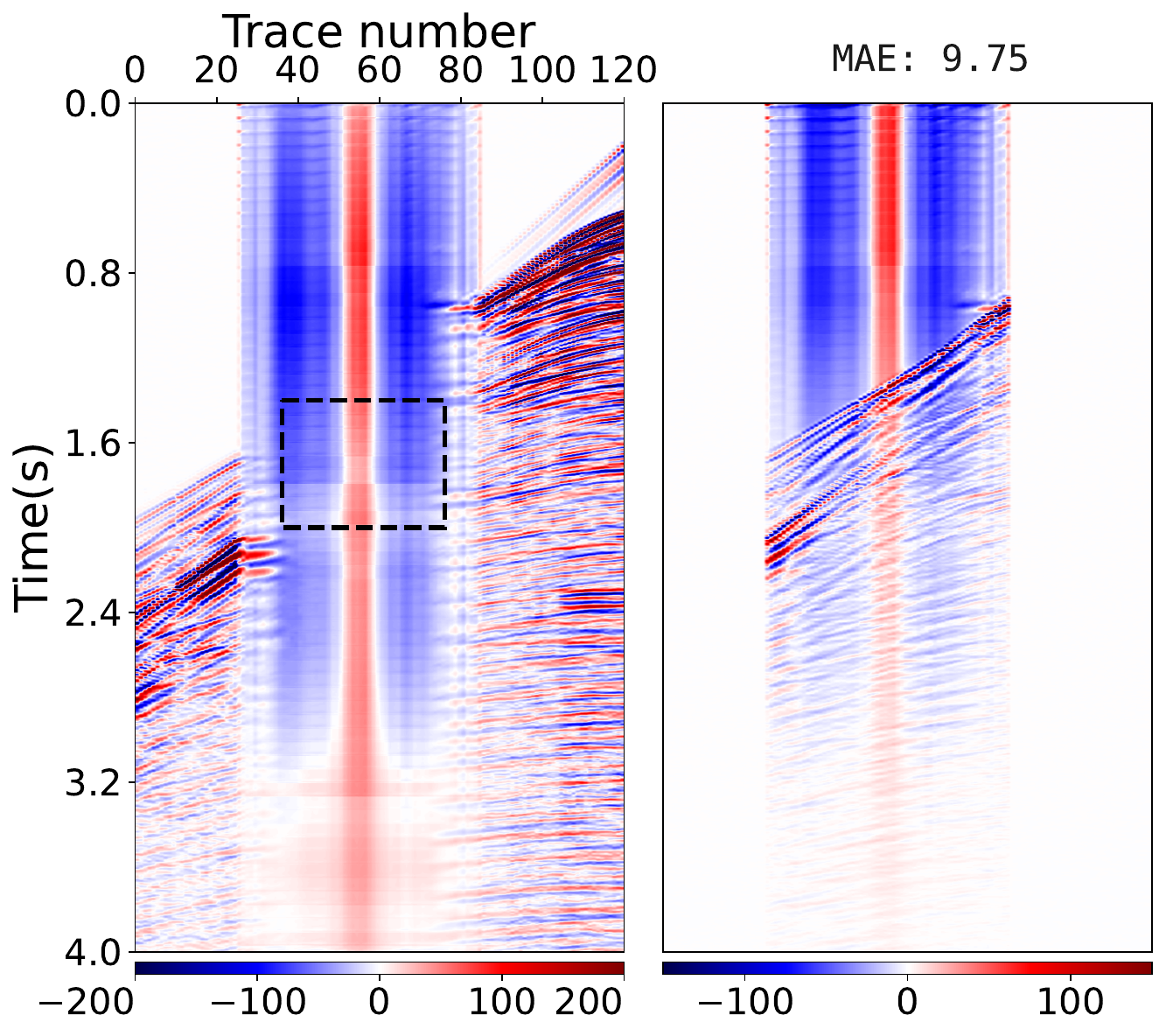}{1}\hfill
			\gatherpanel{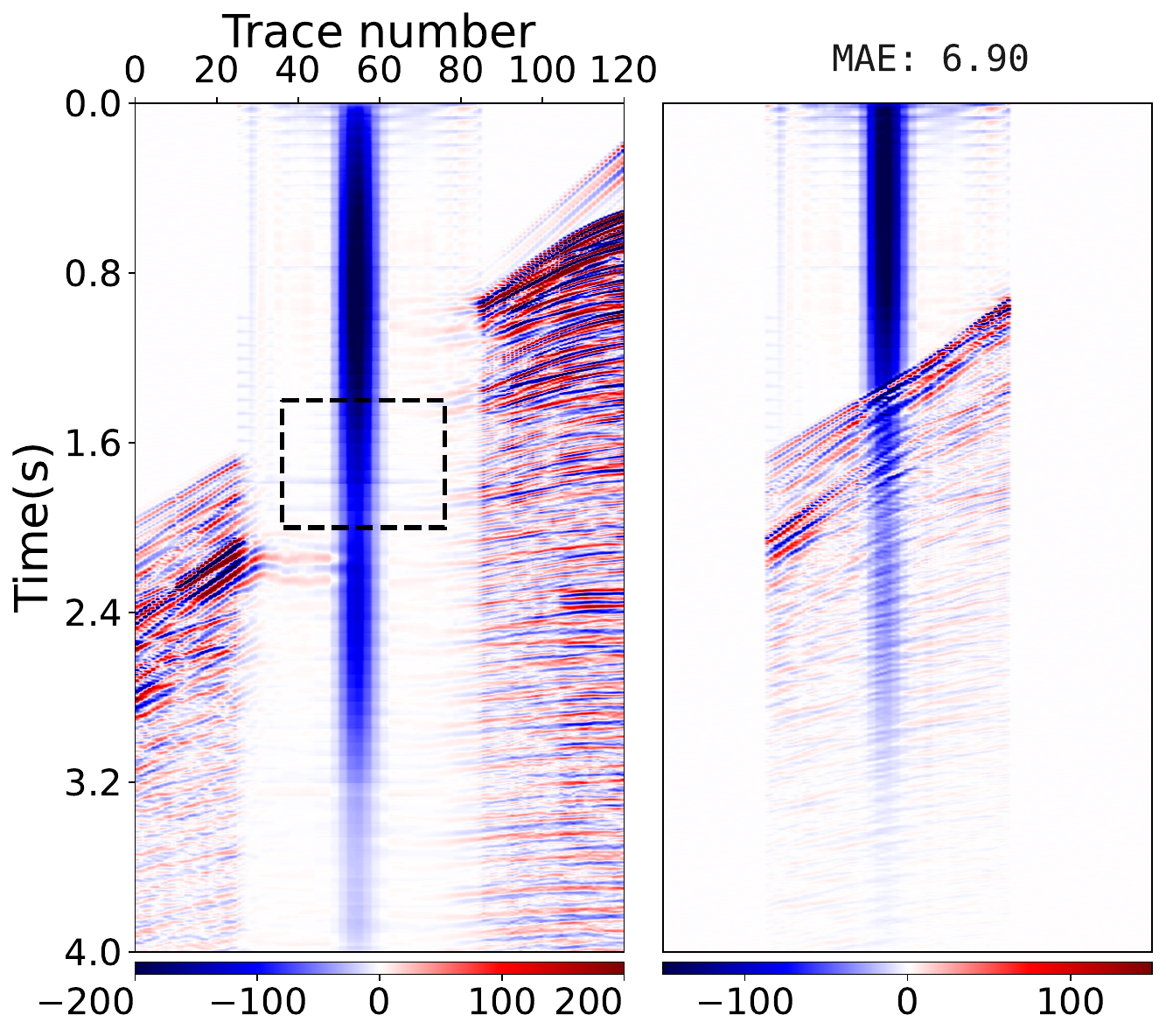}{1}\hfill
			\gatherpanel{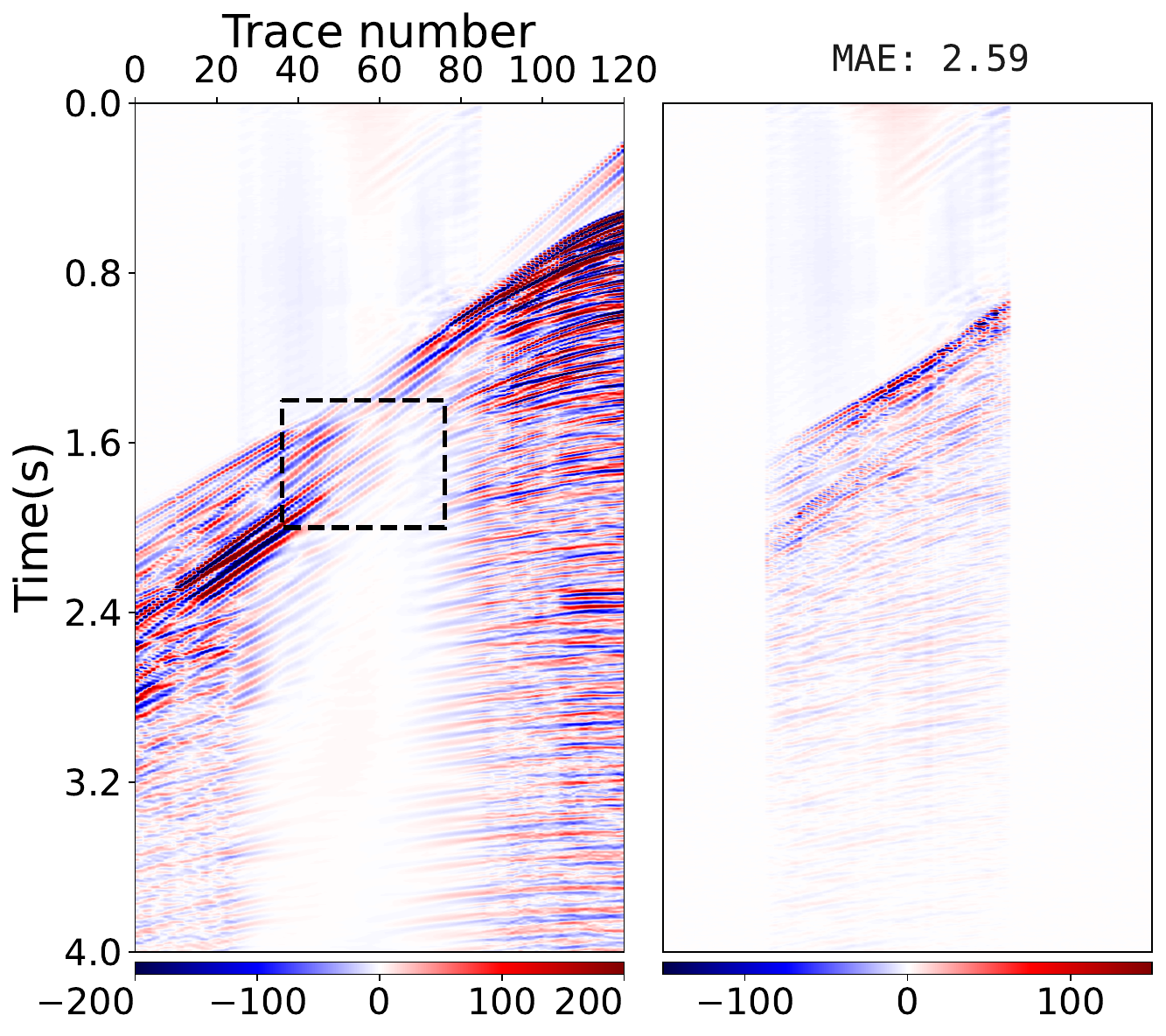}{1}
	\end{minipage}}\par
	%\subfloat[SEG C3, random (section 24\_24)]{\label{fig:segc3_results}\begin{minipage}{\textwidth}\centering
			%\gatherpanel{figures/segc3/shot_24_24_Ori.pdf}{0}\hfill
			%\gatherpanel{figures/segc3/shot_24_24_Miss.pdf}{0}\hfill
			%\gatherpanel{figures/segc3/shot_24_24_Curvelet.pdf}{1}\hfill
			%\gatherpanel{figures/segc3/shot_24_24_POCSNet.pdf}{1}\hfill
			%\gatherpanel{figures/segc3/shot_24_24_SPNet.pdf}{1}\hfill
			%\gatherpanel{figures/segc3/shot_24_24_D3UNet.pdf}{1}\hfill
			%\gatherpanel{figures/segc3/shot_24_24_Ours.pdf}{1}
			%\end{minipage}}\par
			\caption{Qualitative comparison of reconstruction results on the MAVO dataset under 50\% trace removal. (a) Random undersampling (shot 180). (b) Consecutive large gaps (shot 548). We restore the seismic data to its original amplitude range. }
			\label{fig:comparison}
		\end{figure*}

\subsection{Joint Reconstruction and Transform Supervision}
All stages are trained jointly in an end-to-end manner using the final reconstruction:
\begin{equation}
	L=L_1+\lambda_1 L_{\mathrm{SSIM}}+\lambda_2 L_{\mathrm{FFL}}+\lambda_3 L_{\mathrm{sym}}.\label{L_total}
\end{equation}
Here, the pixel-wise $L_1$ norm ensures global amplitude fidelity. To specifically address the structural disruption caused by large gaps, we introduce the structural similarity loss $L_{\mathrm{SSIM}}$~\cite{wang2004imagequa}, which encourages local structural consistency beyond point-wise amplitude fidelity.

Furthermore, to mitigate high-frequency artifacts and spectral leakage in shallow blank zones, the focal frequency loss (FFL)~\cite{2021jiang} dynamically penalizes difficult frequency components. Let $\Delta= \mathcal{F}(\hat x)-\mathcal{F}(x)$ denote the 2-D Fourier spectral error; FFL is formulated as:
\begin{equation}
	L_{FFL}=\frac1{HW}\sum_{u,v}w_{uv}|\Delta_{uv}|^2,
	\quad w_{uv}=\frac{|\Delta_{uv}|}{\max_{u',v'}|\Delta_{u'v'}|}.\label{eq:ffl}
\end{equation}
The adaptive weight $w_{uv}$ focuses the optimization strictly on hard-to-reconstruct spectral discrepancies, rather than uniformly penalizing all frequencies.

Finally, transform consistency regularizes the unthresholded analysis--synthesis pair:
\begin{equation}
	L_{sym}=\sum_{k=1}^{N}\frac{\|\tilde{T}_k(T_k(x^k))-x^k\|_F^2}{HW}.\label{eq:sym}
\end{equation}
This constraint regularizes the transforms as an approximate inverse pair.

\begin{table}[t]
	%\scriptsize
	\caption{Test-set results with 50\% random trace removal. Bold: best.}
	\centering
	\resizebox {1.0\columnwidth}{!}{
		\begin{tabular}{lcccccc}
			\toprule
			\multirow{1}{*}{Dataset} &\multicolumn{3}{c}{SEG C3} &\multicolumn{3}{c}{MAVO}\\
			\cmidrule(lr){2-4} \cmidrule(lr){5-7}
			Model           &MAE$\downarrow$    &SNR$\uparrow$ & SSIM$\uparrow$   &MAE$\downarrow$   &SNR$\uparrow$ & SSIM$\uparrow$ \\
			\midrule
			Curvelet &2.366e-02 &20.131 &0.784 &1.292e-02 &24.600 &0.869 \\
			POCSNet  &1.664e-02 &22.486 &0.890 &9.237e-03 &26.550 &0.929 \\
			SPNet    &1.236e-02 &22.281 &0.883 &9.444e-03 &26.446 &0.926 \\
			D3UNet   &1.366e-02 &24.304 &0.920 &8.138e-03 &27.840 &0.937 \\
			\midrule
			\textbf{Ours}  &\textbf{7.356e-03} &\textbf{31.082} &\textbf{0.974} &\textbf{4.749e-03} &\textbf{33.988} &\textbf{0.973} \\
			\bottomrule
	\end{tabular}}
	\label{tab:tablerandom}
\end{table}
		
\section{Experiments}\label{Experiments}
\subsection{Data and Training Protocol}
We evaluate our method on the synthetic SEG C3 and the North Sea MAVO field datasets, cropped into $128\times128$ and $256\times112$ patches, respectively. Further partition details are provided in~\cite{jiang2026meta}. Code and datasets are publicly available at https://github.com/ProgrammerJXX/Meta-Interpolation. To simulate severe missing scenarios, incomplete inputs are generated by removing 50\% of the traces via either random undersampling or consecutive large gaps.

The SSI-POCS framework is implemented with $N=8$ unrolled stages, a base feature width of 16, and 4 attention heads. Models are optimized using the Adam optimizer with a batch size of 4, an initial learning rate of $10^{-4}$ (decayed via a cosine annealing schedule), and a weight decay of $10^{-4}$. The training spans 100 epochs. Finally, the loss weights in Eq.~\eqref{L_total} are empirically set to $\lambda_1=30$, $\lambda_2=50$, and $\lambda_3=0.05$ based on validation experiments.

\begin{table}[t]
	%\scriptsize
	\caption{Test-set results with 50\% contiguous trace removal.}
	\centering
	\resizebox {1.0\columnwidth}{!}{
		\begin{tabular}{lcccccc}
			\toprule
			\multirow{1}{*}{Dataset} &\multicolumn{3}{c}{SEG C3} &\multicolumn{3}{c}{MAVO}\\
			\cmidrule(lr){2-4} \cmidrule(lr){5-7}
			Model           &MAE$\downarrow$   &SNR$\uparrow$ & SSIM$\uparrow$  &MAE$\downarrow$   &SNR$\uparrow$ & SSIM$\uparrow$  \\
			\midrule
			Curvelet &2.220e-01 &1.0578 &0.517 &2.333e-01 &1.0598 &0.509 \\
			POCSNet  &3.174e-02 &17.810 &0.724 &1.660e-02 &23.178 &0.866 \\
			SPNet    &3.875e-02 &15.578 &0.719 &2.298e-02 &19.110 &0.860 \\
			D3UNet   &2.825e-02 &18.527 &0.729 &1.317e-02 &24.241 &0.870 \\
			\midrule
			\textbf{Ours}  &\textbf{2.415e-02} &\textbf{20.241} &\textbf{0.817} &\textbf{1.111e-02} &\textbf{27.526} &\textbf{0.898} \\
			\bottomrule
	\end{tabular}}
	\label{tab:tablecontinus}
\end{table}

\subsection{Interpolation Across Large Gaps}
We compare the proposed SSI-POCS against Curvelet and three DUNs: POCSNet~\cite{2023Chen}, SPNet~\cite{2024Wu}, and D3UNet~\cite{2026Wang}. Under 50\% random removal (Table~\ref{tab:tablerandom}), SSI-POCS exceeds D3UNet by 6.778 and 6.148~dB on SEG C3 and MAVO, respectively. The upper row of Fig.~\ref{fig:comparison} shows fewer vertical artifacts and more continuous dipping events. 

\begin{figure}[t]
	\centering
	\hspace{-6mm}
	\subfloat[Reference]{\includegraphics[width=.18\textwidth]{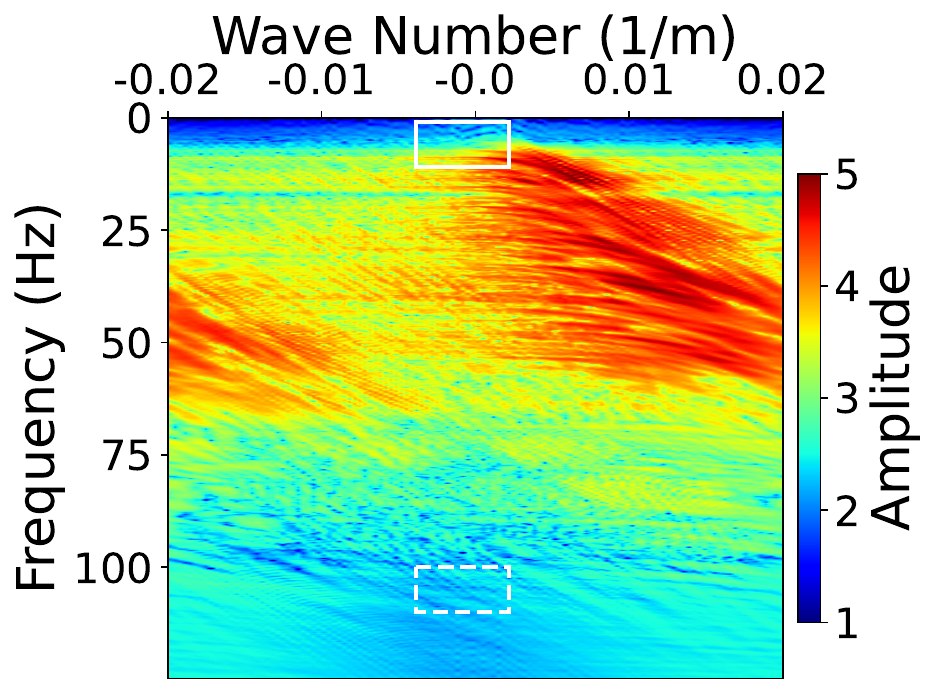}}
	\hspace{-5mm}
	\subfloat[D3UNet]{\includegraphics[width=.18\textwidth]{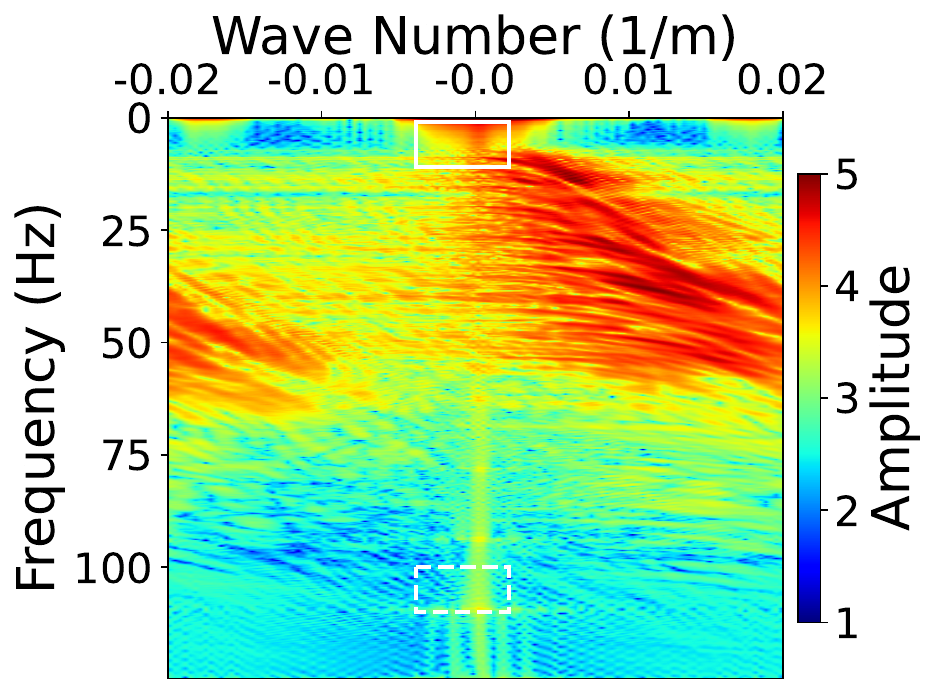}}
	\hspace{-5mm}
	\subfloat[Ours]{\includegraphics[width=.18\textwidth]{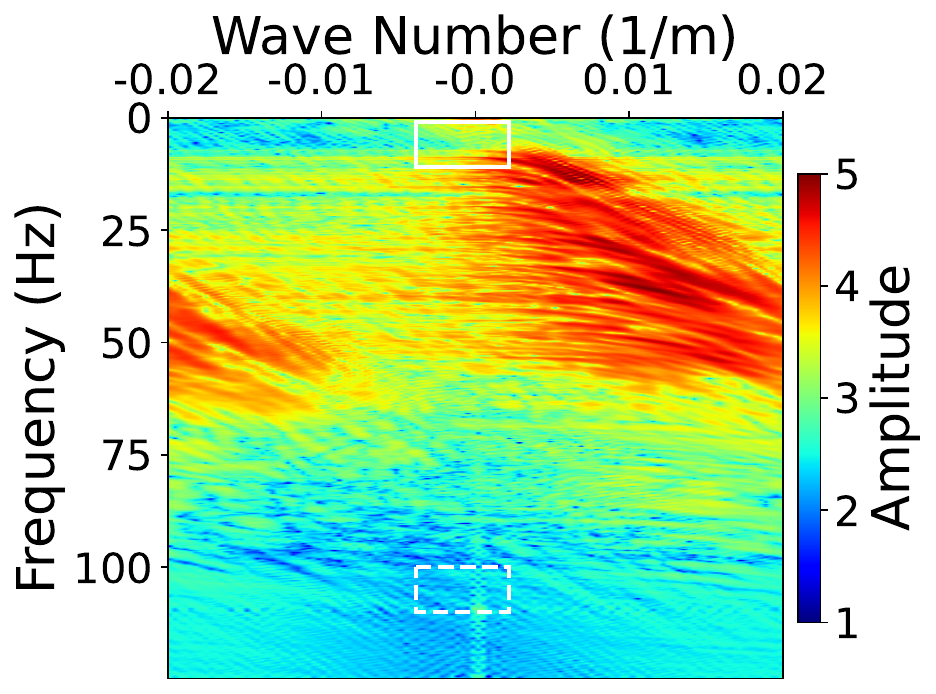}}
	
	\par\smallskip
	
	\hspace{-6mm}
	\subfloat[Reference]{\includegraphics[width=.18\textwidth]{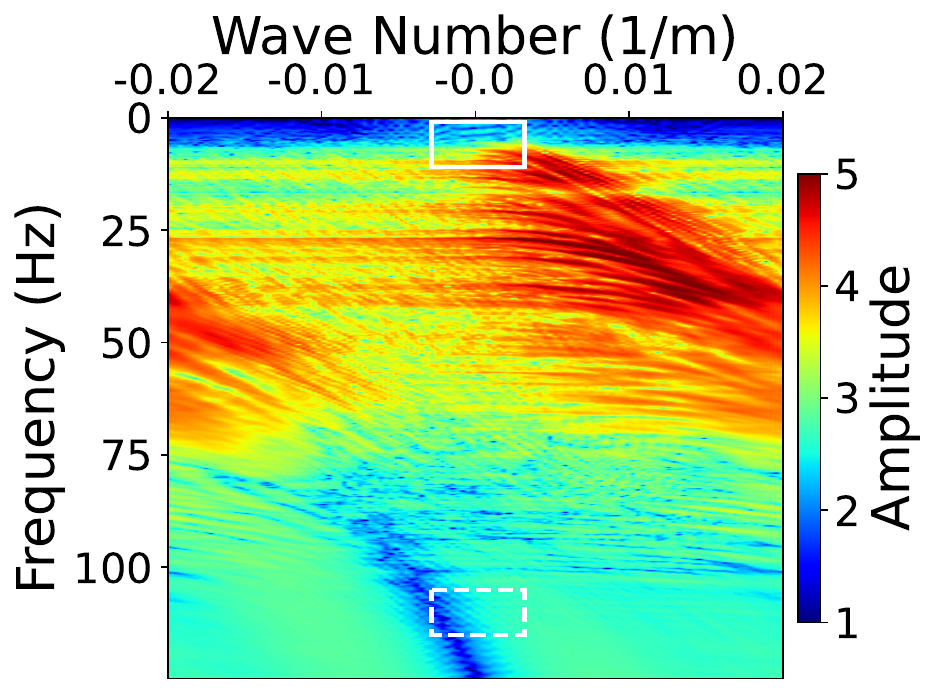}}
	\hspace{-5mm}
	\subfloat[D3UNet]{\includegraphics[width=.18\textwidth]{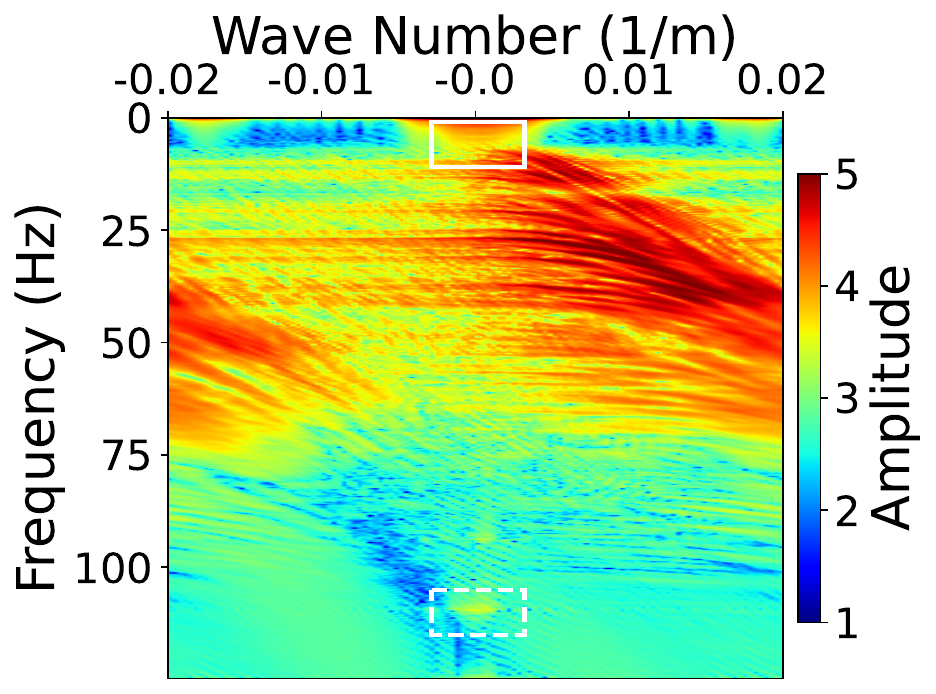}}
	\hspace{-5mm}
	\subfloat[Ours]{\includegraphics[width=.18\textwidth]{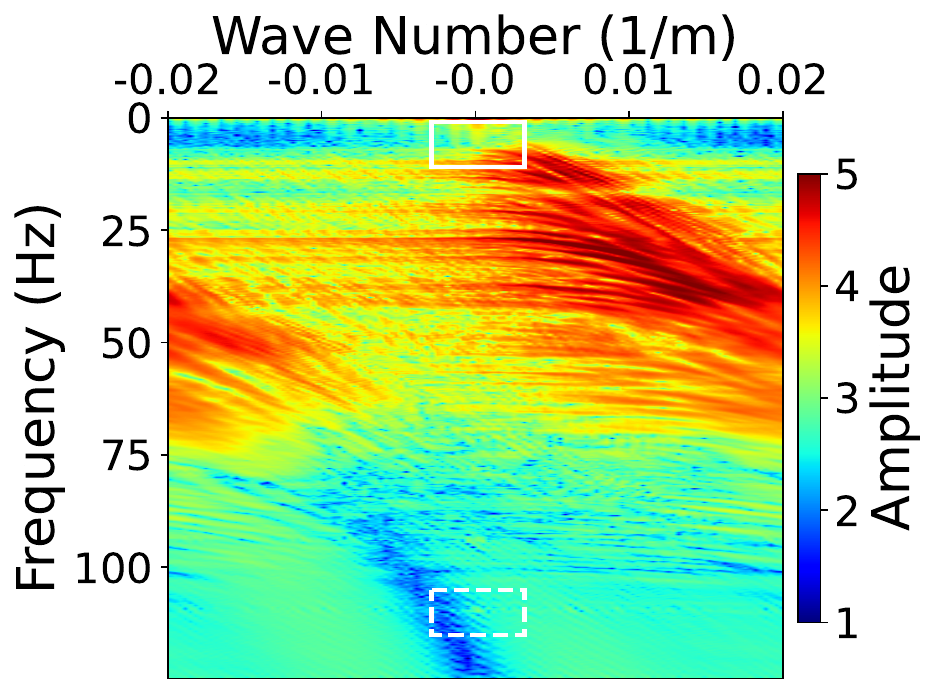}}
	
	\caption{$f$--$k$ spectra for MAVO examples with consecutive trace removal: (a)--(c) 50\% removal (shot 548); (d)--(f) 30\% removal (shot 2).}
	\label{fig:fk_mavo}
\end{figure}

Consecutive trace removal serves as the primary and most stringent benchmark for evaluating reconstruction across large gaps. As detailed in Table~\ref{tab:tablecontinus}, SSI-POCS consistently dominates this challenging task, outperforming D3UNet by 1.714~dB and 3.285~dB in SNR on SEG C3 and MAVO, alongside notable SSIM gains of 0.088 and 0.028. Visually, the lower row of Fig.~\ref{fig:comparison} (MAVO shot 548) shows that Curvelet spreads boundary-induced coefficients across the large gap, producing a saturated negative-amplitude block. Existing DUNs cannot reliably connect both gap boundaries, resulting in vertical artifacts and distorted events. In contrast, SSI-POCS successfully bridges the wide gap, coherently recovering the kinematic trajectories of dipping events in the boxed region. 

The corresponding $f$--$k$ spectra in Fig.~\ref{fig:fk_mavo}(a)--(c) further reveal the spectral effects of the 50\% consecutive gap. D3UNet produces an anomalous vertical ridge near zero wavenumber, including excess energy in the marked low- and high-frequency regions, consistent with its tracewise artifacts. SSI-POCS substantially suppresses this spurious energy and  more closely reproduces the coherent oblique spectral bands of the reference. For 30\% consecutive trace removal, Fig.~\ref{fig:fk_mavo}(d)--(f) likewise shows reduced anomalous energy in the marked regions for SSI-POCS relative to D3UNet. Although deep, highly attenuated reflections near the center of the 50\% gap remain relatively weak due to the severe lack of local information, our method provides more structurally reliable interpolation.

\begin{figure}[t]
	\includegraphics[width=1\columnwidth]{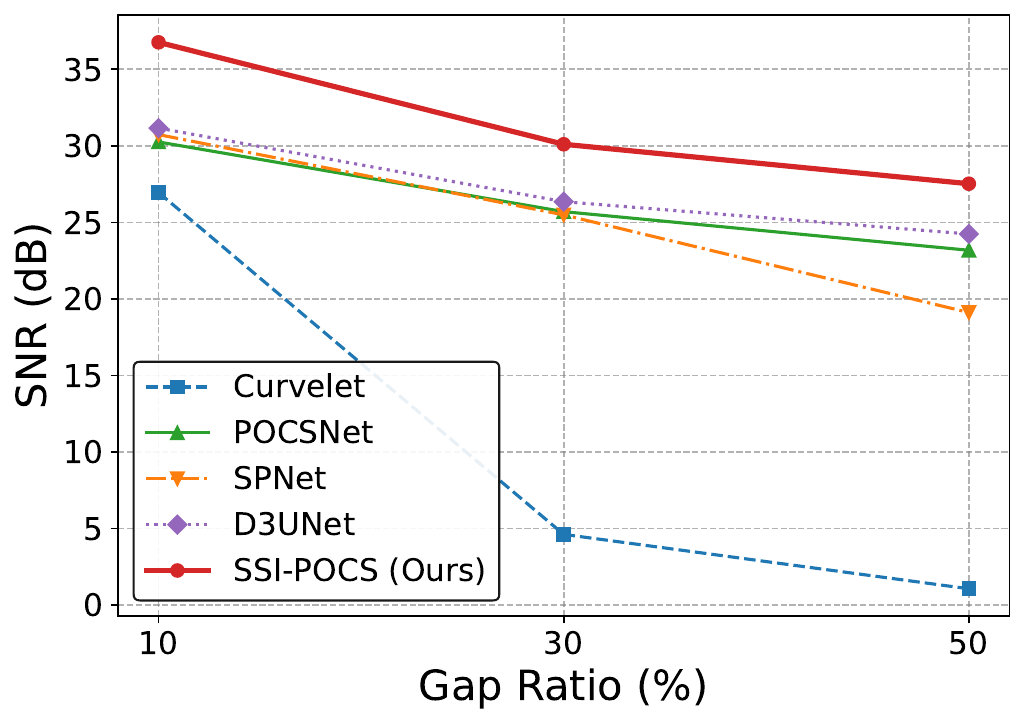}\label{fig:gap_ratio_limit_test_snr}
	\label{fig:gap_limit}
\end{figure}

To evaluate reconstruction capability boundaries, we analyze performance across consecutive missing ratios (10\% to 50\%) on the MAVO dataset. As shown in Fig.~\ref{fig:gap_limit}, Curvelet fails catastrophically beyond 30\% missing, as the gap surpasses the effective support of its transform atoms. While existing DUNs (including the dual-domain D3UNet) show better resilience, their performance still degrades sharply under extreme gaps. This occurs because localized convolutions and finite-support wavelets lack the global receptive field required for long-range extrapolation. In contrast, SSI-POCS degrades gracefully, maintaining a superior SNR of 27.526 dB even at 50\% removal—a performance comparable to baselines operating at merely 10\% to 30\% missing ratios. This confirms that our global Fourier spectral prior effectively bridges extensive spatial gaps.

\begin{figure}[t]
	\centering
	\subfloat[Reference]{\includegraphics[width=.18\textwidth]{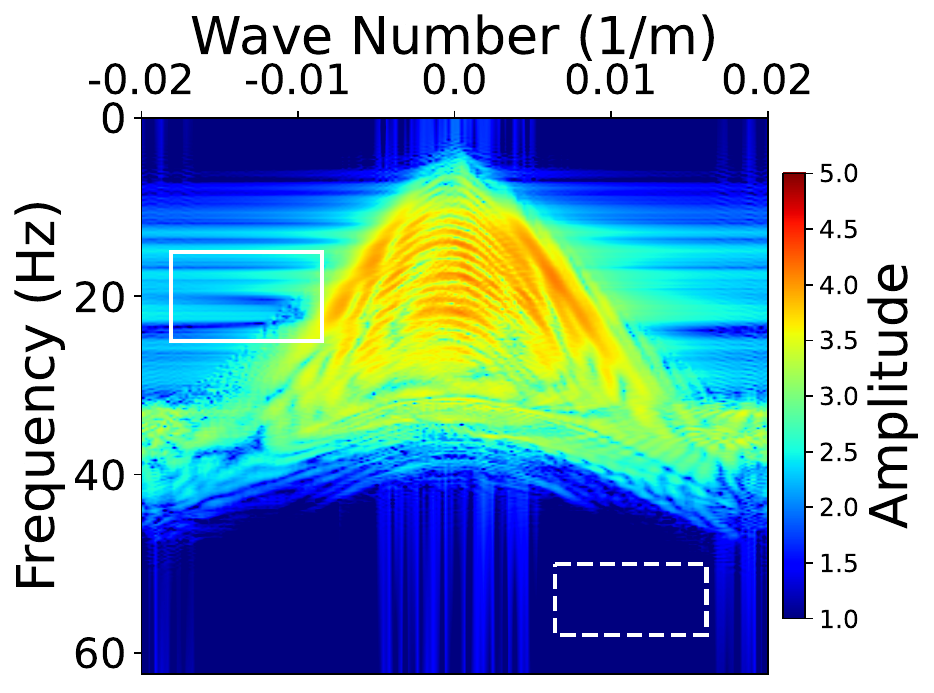}}
	\subfloat[Missing]{\includegraphics[width=.18\textwidth]{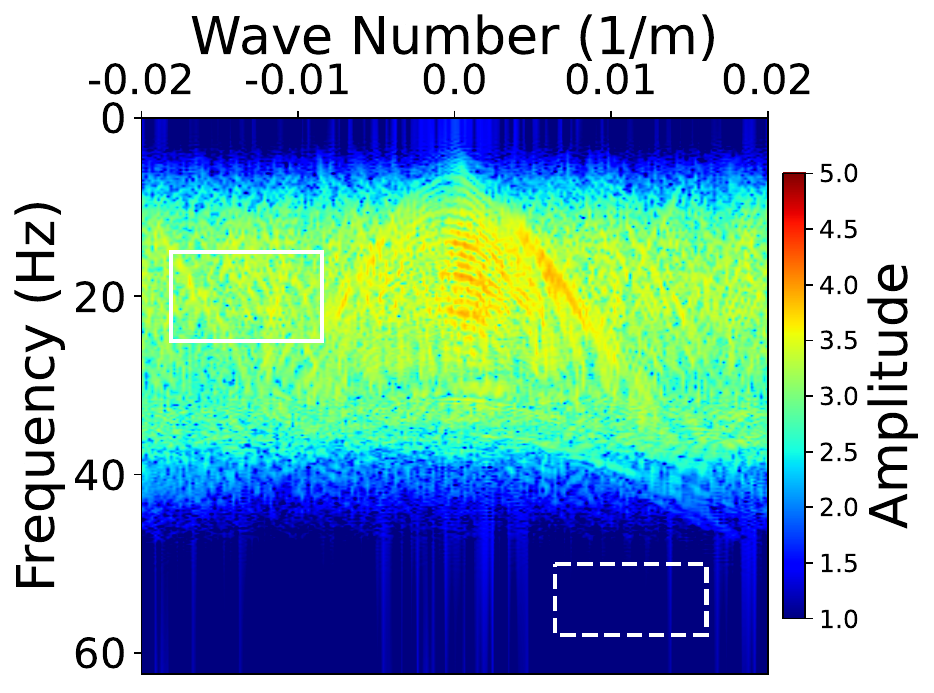}}\hfill
	\subfloat[Spa--Spa]{\includegraphics[width=.18\textwidth]{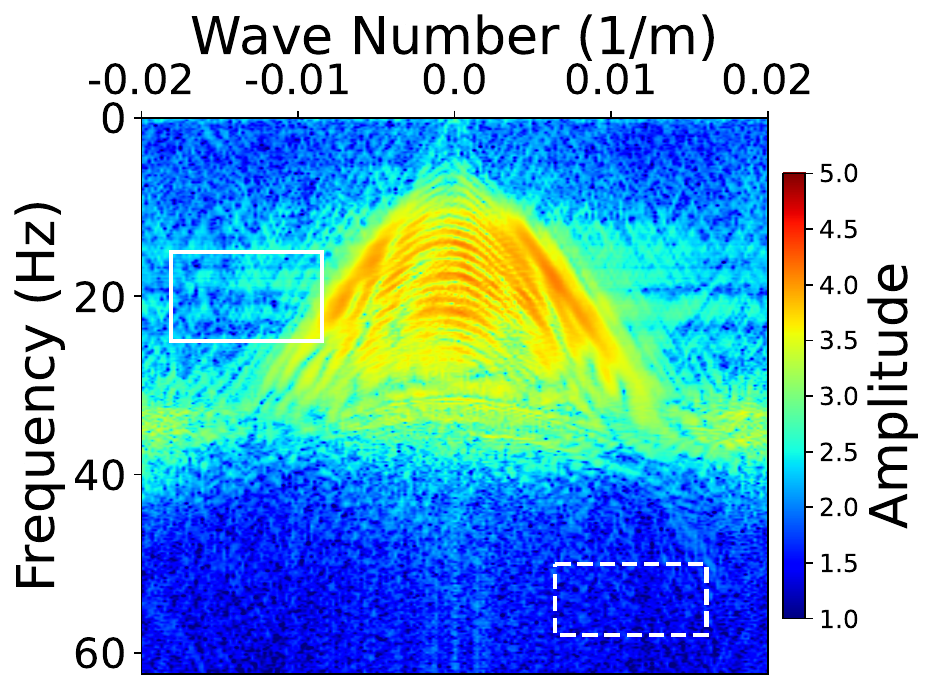}}
	\subfloat[Spa--FFC]{\includegraphics[width=.18\textwidth]{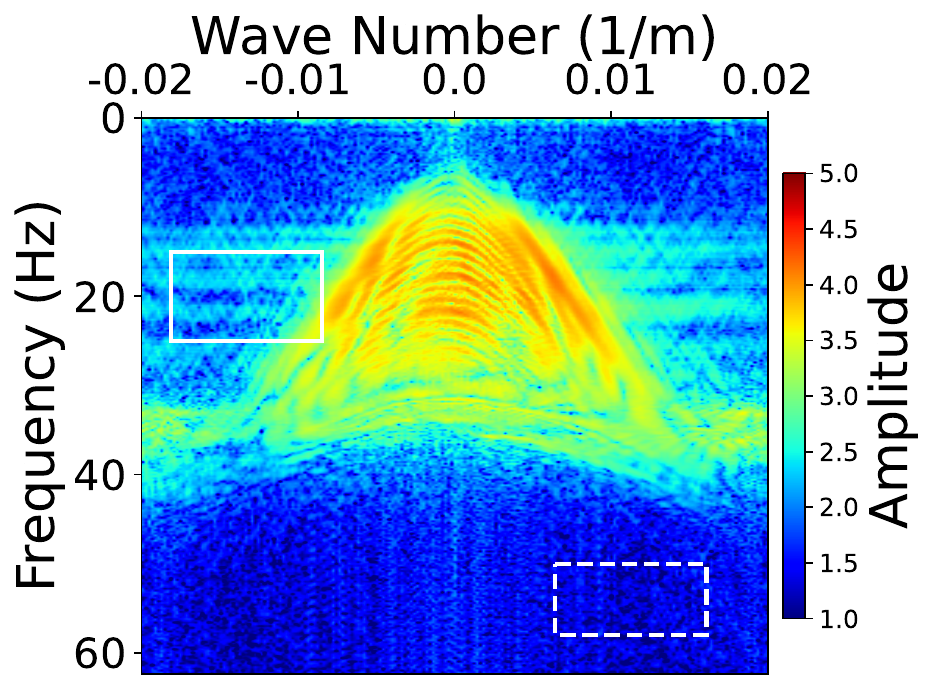}}
	\caption{Branch ablation in the $f$--$k$ domain (SEG C3 section 24-24).}
	\label{fig:fk_ffc_ablation}
\end{figure}

\subsection{Architecture and Loss Ablations}\label{sec:ablation}
To validate the architectural design, we conduct ablation studies using 30\% of the SEG C3 shots under 50\% random trace undersampling (Table~\ref{tab:ablation_all}). Compared to a homogeneous Spa--Spa baseline, the heterogeneous Spa--FFC prior better preserves principal coherent bands and successfully suppresses high-frequency background noise in the $f-k$ domain (Fig.~\ref{fig:fk_ffc_ablation}). Furthermore, upgrading the naive concatenation to our DFM yields a notable SNR improvement of 0.567~dB and reduces the MSE by 11.6\%, confirming the necessity of explicit bidirectional cross-domain interaction.

\begin{figure}[t]
	\centering
	%\subfloat[Reference]{\includegraphics[width=.24\columnwidth,height=1.2in]{figures/ablation/imputed_data_shot_43_92_Ori.pdf}}\hfill
	%\subfloat[Missing]{\includegraphics[width=.24\columnwidth,height=1.2in]{figures/ablation/imputed_data_shot_43_92_Miss.pdf}}\hfill
	\subfloat[Without SSIM]{\includegraphics[width=.19\textwidth]{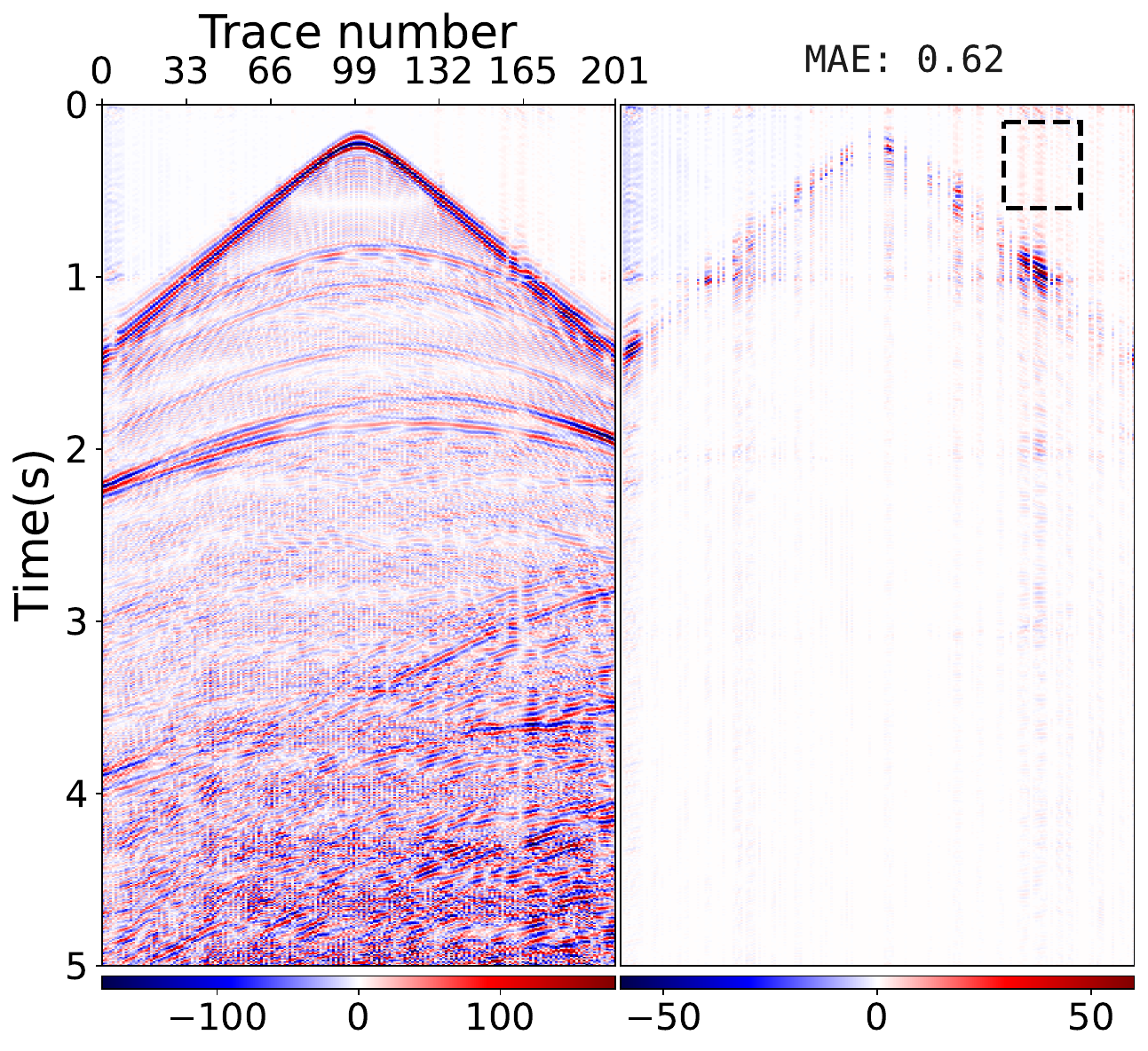}}
	\subfloat[With SSIM]{\includegraphics[width=.19\textwidth]{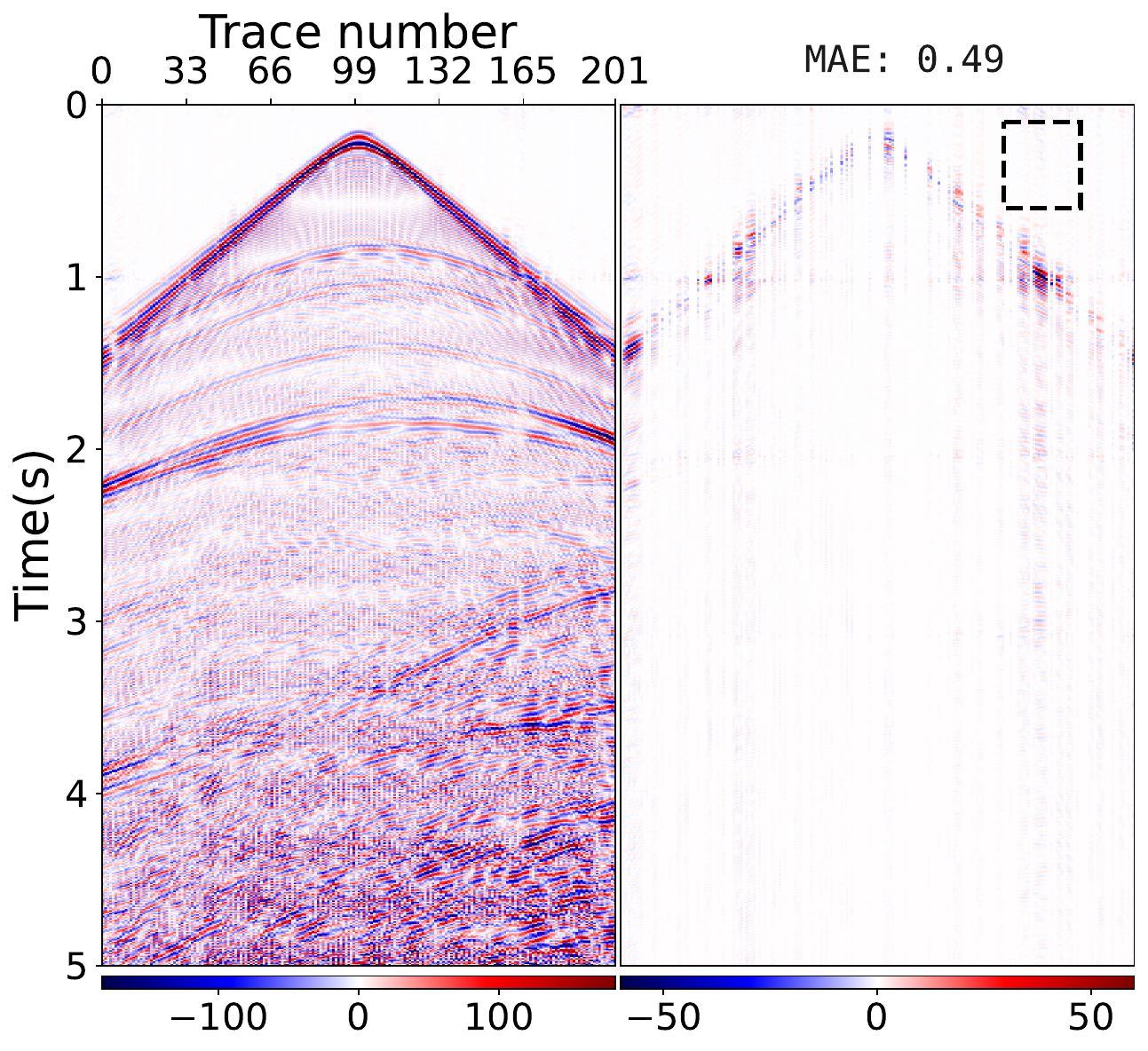}}\hfill
	\subfloat[Without FFL]{\includegraphics[width=.19\textwidth]{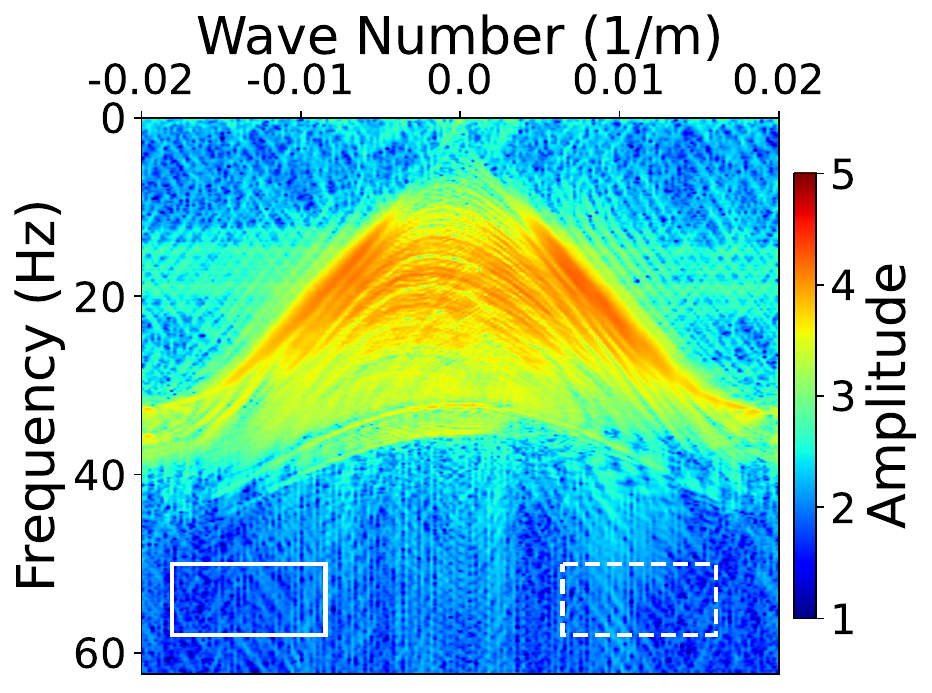}}
	\subfloat[With FFL]{\includegraphics[width=.19\textwidth]{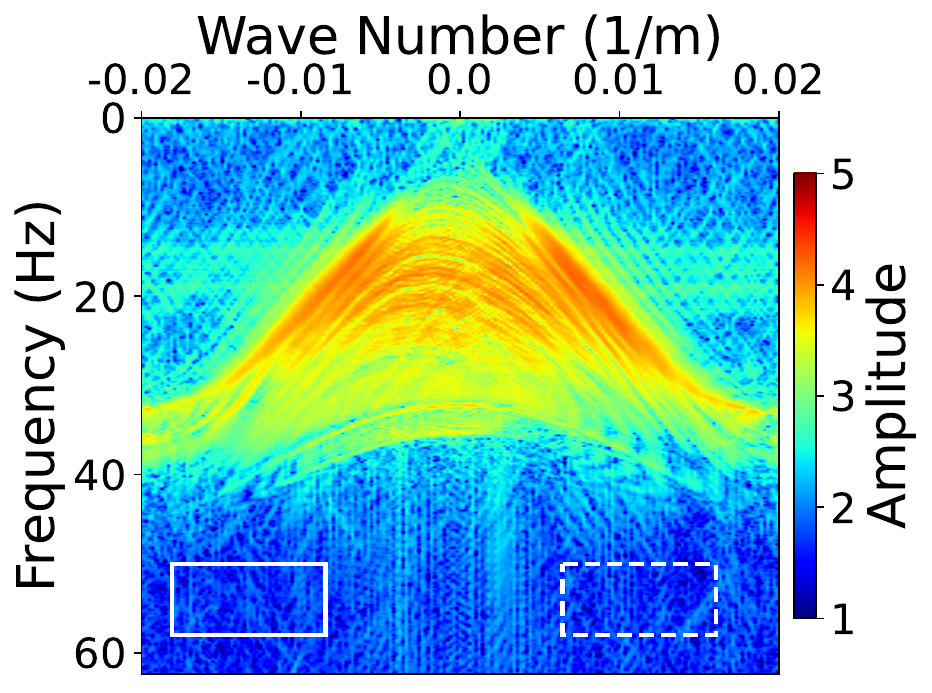}}
	
	\caption{Loss ablation on SEG C3. (a) and (b) Effect of $L_{\mathrm{SSIM}}$ (highlighting complex dipping events). (c) and (d) Effect of $L_{\mathrm{FFL}}$ on the $f$--$k$ spectra (highlighting spectral leakage).}
	\label{fig:SSIM_fk_ffl_ablation}
\end{figure}

We further progressively evaluate the multi-dimensional supervision. Incorporating $L_{\mathrm{SSIM}}$ into the baseline ($L_1+L_{\mathrm{sym}}$) yields a 0.231~dB SNR gain. Qualitatively, Fig.~\ref{fig:SSIM_fk_ffl_ablation}(a)--(b) demonstrates that SSIM effectively weakens tracewise residuals near shallow events, reducing the section-level mean absolute error (MAE) from 0.62 to 0.49. Although the subsequent addition of $L_{\mathrm{FFL}}$ provides a marginal numerical SNR gain (as standard SNR metrics are heavily dominated by high-amplitude, low-frequency components), its spectral contribution is visually significant. As highlighted in Fig.~\ref{fig:SSIM_fk_ffl_ablation}(c)--(d), FFL successfully eliminates incoherent diffuse energy in high-frequency regions while strictly preserving essential spectral bands. These results support the complementary roles of structural and spectral supervision.

\begin{table}[t]
	\centering
	\caption{Comprehensive Ablation Study.}
	\label{tab:ablation_all}
	\resizebox{\columnwidth}{!}{
	\begin{tabular}{cc|cc|ccc}\toprule
		\multicolumn{2}{c|}{Fusion} & \multicolumn{2}{c|}{Loss Functions} & \multirow{2}{*}{MSE$\downarrow$} & \multirow{2}{*}{SSIM$\uparrow$} & \multirow{2}{*}{SNR$\uparrow$} \\
		Concat & DFM & $L_{\mathrm{SSIM}}$ & $L_{\mathrm{FFL}}$ & & & \\\midrule
		$\checkmark$ & & & & 5.604e-04 & 0.954 & 26.936 \\
		& $\checkmark$ & & & 4.952e-04 & 0.958 & 27.503 \\
		& $\checkmark$ & $\checkmark$ & & 4.723e-04 & \textbf{0.963} & 27.734 \\
		& $\checkmark$ & $\checkmark$ & $\checkmark$ & \textbf{4.719e-04} & \textbf{0.963} & \textbf{27.736} \\\bottomrule
		\multicolumn{7}{l}{\footnotesize *All models use the Spa--FFC architecture and the base $L_1+SymLoss$ supervision.}
	\end{tabular}
}
\end{table}

\begin{figure}[t]
	\subfloat[]{\includegraphics[height=1in]{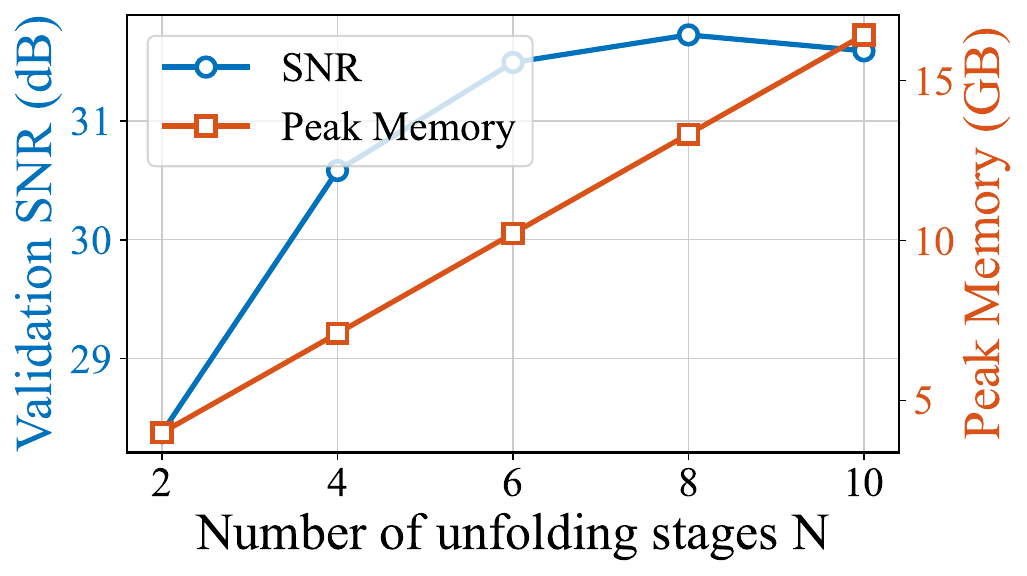}\label{fig:random_0.5_T_vs_SNR_CC}}
	\subfloat[]{\includegraphics[height=1in]{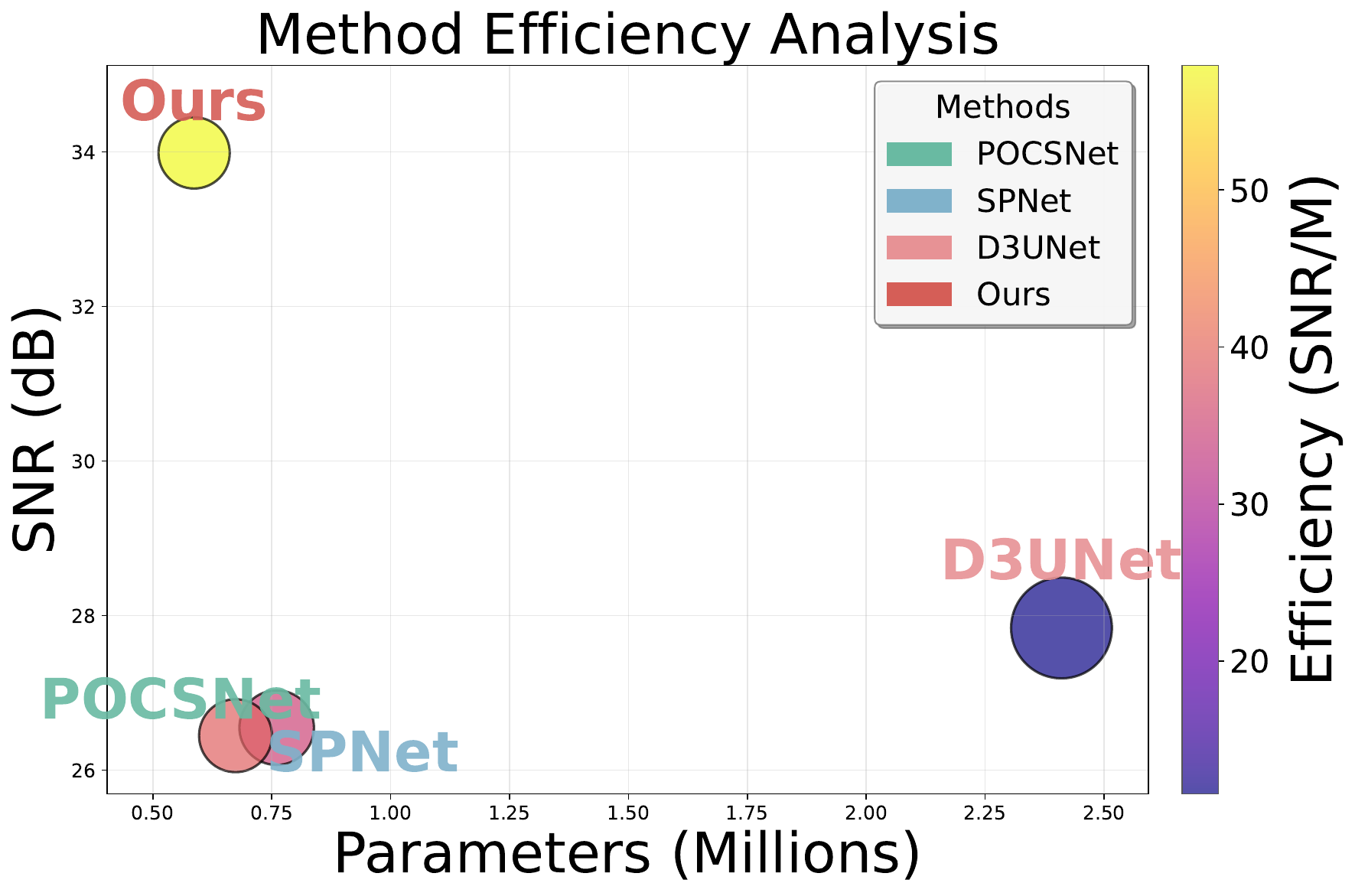}\label{fig:performance_efficiency}}
	\caption{Model performance and complexity analysis. (a) Impact of the unrolling depth $N$ on validation SNR and computational memory. (b) Performance-complexity trade-off: SNR versus parameter count.}
	\label{fig:model performance}
\end{figure}
\subsection{Unfolding Depth and Model Efficiency}
Finally, Fig.~\ref{fig:model performance}(a) evaluates the unrolling depth $N$ on MAVO (50\% random missing). Validation SNR peaks at 31.726~dB at $N=8$ (up from 28.375~dB at $N=2$). Deploying $N=10$ incurs a 23.3\% memory penalty and degrades SNR to 31.592~dB, likely due to overfitting in overly deep architectures, establishing $N=8$ as optimal. Moreover, Fig.~\ref{fig:model performance}(b) demonstrates a better performance-complexity trade-off. SSI-POCS achieves a 33.988~dB SNR with only 0.587~M parameters, significantly outperforming D3UNet ($\sim$27.840~dB, 2.411~M). This 75.66\% reduction in parameter count indicates that the proposed dual-domain formulation achieves competitive reconstruction accuracy with a substantially smaller model.

\section{Conclusion}\label{Conclusion}
SSI-POCS effectively mitigates the severe ill-posedness of interpolating large consecutive gaps. Outperforming traditional deep unrolling networks, it substantially improves event continuity across extensive missing regions and suppresses high-frequency artifacts. Notably, it improves reconstruction accuracy and event continuity while maintaining a compact parameter footprint. However, reconstructing highly attenuated reflections at the absolute center of wide gaps remains challenging due to the fundamental absence of local structural information. Future work will address this by extending the framework to 3D seismic geometries to exploit crossline coherence, and incorporating unsupervised domain adaptation to better generalize to complex field data.

% References are embedded to keep this revision self-contained: reference.bib
% is left unchanged (it contains duplicate keys and malformed entries).
\bibliographystyle{IEEEtran}
\bibliography{SSIPOCS}

\begin{thebibliography}{10}
\providecommand{\url}[1]{#1}
\csname url@samestyle\endcsname
\providecommand{\newblock}{\relax}
\providecommand{\bibinfo}[2]{#2}
\providecommand{\BIBentrySTDinterwordspacing}{\spaceskip=0pt\relax}
\providecommand{\BIBentryALTinterwordstretchfactor}{4}
\providecommand{\BIBentryALTinterwordspacing}{\spaceskip=\fontdimen2\font plus
\BIBentryALTinterwordstretchfactor\fontdimen3\font minus
  \fontdimen4\font\relax}
\providecommand{\BIBforeignlanguage}[2]{{%
\expandafter\ifx\csname l@#1\endcsname\relax
\typeout{** WARNING: IEEEtran.bst: No hyphenation pattern has been}%
\typeout{** loaded for the language `#1'. Using the pattern for}%
\typeout{** the default language instead.}%
\else
\language=\csname l@#1\endcsname
\fi
#2}}
\providecommand{\BIBdecl}{\relax}
\BIBdecl

\bibitem{1987Ronen}
J.~Ronen, ``Wave-equation trace interpolation,'' \emph{Geophysics}, vol.~52,
  no.~7, pp. 973--984, Jul. 1987.

\bibitem{1991spitz}
S.~Spitz, ``Seismic trace interpolation in the fx domain,'' \emph{Geophysics},
  vol.~56, no.~6, pp. 785--794, 1991.

\bibitem{2007Zwartjes}
P.~Zwartjes and M.~Sacchi, ``Fourier reconstruction of nonuniformly sampled,
  aliased seismic data,'' \emph{Geophysics}, vol.~72, no.~1, pp. V21--V32,
  2007.

\bibitem{2017Zhang}
H.~Zhang, X.-H. Chen, and L.-Y. Zhang, ``3d simultaneous seismic data
  reconstruction and noise suppression based on the curvelet transform,''
  \emph{Applied Geophysics}, vol.~14, no.~1, pp. 87--95, 2017.

\bibitem{2018Gong}
X.~Gong, S.~Wang, and L.~Du, ``Seismic data reconstruction using a
  sparsity-promoting apex shifted hyperbolic radon-curvelet transform,''
  \emph{Studia Geophysica et Geodaetica}, vol.~62, no.~3, pp. 450--465, 2018.

\bibitem{Hennenfent2008Gap}
G.~Hennenfent and F.~J. Herrmann, ``Simply denoise: Wavefield reconstruction
  via jittered undersampling,'' \emph{Geophysics}, vol.~73, no.~3, pp.
  V19--V28, 2008.

\bibitem{Yang2012Curvelet}
P.~Yang, J.~Gao, and W.~Chen, ``Curvelet-based {POCS} interpolation of
  nonuniformly sampled seismic records,'' \emph{Journal of Applied Geophysics},
  vol.~79, pp. 90--99, 2012.

\bibitem{Yu2022Attention}
J.~Yu and B.~Wu, ``Attention and hybrid loss guided deep learning for
  consecutively missing seismic data reconstruction,'' \emph{IEEE Transactions
  on Geoscience and Remote Sensing}, vol.~60, pp. 1--8, 2021.

\bibitem{Huang2022SelfSupervised}
H.~Huang, T.~Wang, J.~Cheng, Y.~Xiong, C.~Wang, and J.~Geng, ``Self-supervised
  deep learning to reconstruct seismic data with consecutively missing
  traces,'' \emph{IEEE Transactions on Geoscience and Remote Sensing}, vol.~60,
  pp. 1--14, 2022.

\bibitem{Wei2022Hybrid}
X.~Wei, C.~Zhang, H.~Wang, Z.~Zhao, D.~Xiong, S.~Xu, J.~Zhang, and S.-W. Kim,
  ``Hybrid loss-guided coarse-to-fine model for seismic data consecutively
  missing trace reconstruction,'' \emph{IEEE Transactions on Geoscience and
  Remote Sensing}, vol. 60. ~\url{https://doi.org/10.1109/TGRS.2022.3223421},
  no. 5923315, 2022.

\bibitem{Cheng2024SIT}
M.~Cheng, J.~Lin, X.~Dong, S.~Lu, and T.~Zhong, ``Seismic interpolation
  transformer for consecutively missing data: A case study in {DAS-VSP} data,''
  \emph{IEEE Transactions on Geoscience and Remote Sensing}, vol.~62, pp.
  1--12, 2024.

\bibitem{Tian2025FRUNet}
Y.~Tian, L.~Fu, W.~Fang, and T.~Li, ``{FR-UNet}: A feature restoration-based
  {UNet} for seismic data consecutively missing trace interpolation,''
  \emph{IEEE Transactions on Geoscience and Remote Sensing}, vol.~63, pp.
  1--10, 2025.

\bibitem{2023Chen}
Y.~Chen, S.~Yu, and J.~Ma, ``A projection-onto-convex-sets network for 3d
  seismic data interpolation,'' \emph{Geophysics}, vol.~88, no.~3, pp.
  V249--V265, MAY-JUN 2023.

\bibitem{2024Wu}
M.~Wu, L.~Fu, W.~Fang, and J.~Cao, ``Sparse prior-net: A sparse prior-based
  deep network for seismic data interpolation,'' \emph{Geophysics}, vol.~89,
  no.~1, pp. V37--V47, 2024.

\bibitem{2026Wang}
Z.~Wang, X.~Gong, Z.~Ma, G.~Peng, and Y.~Long, ``Dual-domain deep unfolding
  network with implicit regularization for seismic data reconstruction,''
  \emph{IEEE Transactions on Geoscience and Remote Sensing}, vol.~64, 2026.

\bibitem{wang2004imagequa}
Z.~Wang, A.~C. Bovik, H.~R. Sheikh, and E.~P. Simoncelli, ``Image quality
  assessment: from error visibility to structural similarity,'' \emph{IEEE
  transactions on image processing}, vol.~13, no.~4, pp. 600--612, 2004.

\bibitem{2021jiang}
L.~Jiang, B.~Dai, W.~Wu, and C.~C. Loy, ``Focal frequency loss for image
  reconstruction and synthesis,'' in \emph{2021 IEEE/CVF International
  Conference on Computer Vision (ICCV)}.\hskip 1em plus 0.5em minus 0.4em\relax
  IEEE, 2021, pp. 13\,899--13\,909.

\bibitem{2022Restormer}
S.~W. Zamir, A.~Arora, S.~Khan, M.~Hayat, F.~S. Khan, and M.-H. Yang,
  ``Restormer: Efficient transformer for high-resolution image restoration,''
  in \emph{Proceedings of the IEEE/CVF conference on computer vision and
  pattern recognition}, 2022, pp. 5728--5739.

\bibitem{chi2020fast}
L.~Chi, B.~Jiang, and Y.~Mu, ``Fast fourier convolution,'' \emph{Advances in
  Neural Information Processing Systems}, vol.~33, pp. 4479--4488, 2020.

\bibitem{2025bai}
H.~Bai, Z.~Zhao, J.~Zhang, Y.~Wu, L.~Deng, Y.~Cui, B.~Jiang, and S.~Xu,
  ``Refusion: Learning image fusion from reconstruction with learnable loss via
  meta-learning,'' \emph{International Journal of Computer Vision}, vol. 133,
  no.~5, pp. 2547--2567, 2025.

\bibitem{jiang2026meta}
B.~Jiang, C.~Zhang, X.~Wei, A.~Di, N.~Zhao, and D.~Xiong, ``Meta-interpolation:
  An efficient seismic data interpolation framework for adaptive spatial
  continuity modelling,'' \emph{Geophysical Prospecting}, vol.~74, no.~7, p.
  e70252, 2026.

\end{thebibliography}
\end{document}